\pdfoutput=1
\documentclass{article}
\ifdefined\pdfsuppresswarningdupdest
\fi

\usepackage[main, final]{neurips_2026}

\usepackage[utf8]{inputenc} %
\usepackage[T1]{fontenc} %
\usepackage[dvipsnames]{xcolor} %
\usepackage{hyperref}

\usepackage{url} %
\usepackage{booktabs} %
\usepackage{amsfonts} %
\usepackage{nicefrac} %
\usepackage{microtype} %
\usepackage{float}
\usepackage{placeins} %
\usepackage{amsmath,amssymb}
\usepackage{enumitem}
\usepackage{algorithm, algpseudocode}
\usepackage{graphicx}

\title{Don't Discard Your Rollouts: Reusing Teacher RL Traces for Student Distillation}

\author{%
  Vashisth Tiwari\thanks{Correspondence to \texttt{vashistt@cs.cmu.edu}}\\
  Carnegie Mellon University \\
\And
  Emma Strubell\thanks{Equal advising.}\\
  Carnegie Mellon University
\And
  J. Zico Kolter\textsuperscript{$\dagger$}\\
  Carnegie Mellon University
}

\definecolor{paperboxframe}{HTML}{C9BEDD}
\definecolor{paperboxfill}{HTML}{FBF9FD}
\newsavebox{\stylehypbox}
\newenvironment{paperbox}[1]
  {\par\smallskip\noindent\begin{lrbox}{\stylehypbox}\begin{minipage}{\dimexpr\linewidth-2\fboxsep-2\fboxrule\relax}\ifx\relax#1\relax\else\textbf{#1}\ \fi}
  {\end{minipage}\end{lrbox}\fcolorbox{paperboxframe}{paperboxfill}{\usebox{\stylehypbox}}\par\smallskip}
\newenvironment{stylehypothesis}[1]{\begin{paperbox}{#1}\itshape}{\end{paperbox}}

\newcommand{\Drl}{\ensuremath{\mathcal{D}_{\mathrm{RL}}}}
\newcommand{\Dpool}{\ensuremath{\mathcal{D}_{\mathrm{rollouts}}}}
\newcommand{\Dsynth}{\ensuremath{\mathcal{D}_{\mathrm{synthetic}}}}

\newcommand{\DsftRand}{\ensuremath{\mathcal{D}_{\mathrm{SFT}}^{\mathrm{random}}}}
\newcommand{\DsftCurr}{\ensuremath{\mathcal{D}_{\mathrm{SFT}}^{\mathrm{curr,top1}}}}

\newcommand{\DsftAllp}{\ensuremath{\mathcal{D}_{\mathrm{SFT}}^{\mathrm{allp,top1}}}}
\newcommand{\DdpoRand}{\ensuremath{\mathcal{D}_{\mathrm{DPO}}^{\mathrm{random}}}}
\newcommand{\DdpoCurr}{\ensuremath{\mathcal{D}_{\mathrm{DPO}}^{\mathrm{curr,top1}}}}
\newcommand{\DtwoStage}{\ensuremath{\DsftAllp\!\to\!\DdpoCurr}}

\definecolor{rolloutsaved}{RGB}{204,102,0}

\definecolor{hlprompt}{RGB}{255,245,197}
\definecolor{hlcompl}{RGB}{219,244,225}
\definecolor{hlreward}{RGB}{255,231,207}
\definecolor{hllogp}{RGB}{255,223,237}
\definecolor{hlpass}{RGB}{223,236,255}
\newcommand{\hlm}[2]{{\setlength{\fboxsep}{0.8pt}\mathchoice
  {\colorbox{#1}{$\displaystyle #2$}}{\colorbox{#1}{$\textstyle #2$}}%
  {\colorbox{#1}{$\scriptstyle #2$}}{\colorbox{#1}{$\scriptscriptstyle #2$}}}}
\newcommand{\hlx}[1]{#1} %
\newcommand{\hly}[1]{#1}
\newcommand{\hlr}[1]{#1}
\newcommand{\hlpi}[1]{\hlm{hllogp}{#1}}
\newcommand{\hlmu}[1]{\hlm{hlpass}{#1}}


\begin{document}

\maketitle

\begin{abstract}
    RL post-training methods such as PPO and GRPO are computationally expensive, with substantial resources devoted to sampling rollouts, which contain multiple completions per prompt along with their respective rewards. Canonical distillation, however, discards these traces and uses more compute to regenerate data from the final teacher. We ask if and how we can reuse the saved teacher RL rollouts to distil student models. Reusing rollouts is non-trivial: naive SFT on the rollouts underperforms canonical distillation, and naive DPO, while performant, is unstable, collapsing on weaker students. We show that treating rollout reuse as a \textit{data selection problem} using metadata already recorded during GRPO substantially reduces this gap. We use prompt pass rates to order the data, using the easiest prompts for an SFT warm start prior to using mixed-success prompts for DPO. For each prompt, we use saved teacher log-likelihoods as a proxy for selecting student-compatible completions. Across mathematical reasoning and Python coding tasks, with students from multiple model families, our two-stage recipe achieves comparable average performance to canonical distillation, while incurring no additional inference cost for dataset construction.
\end{abstract}
\section{Introduction}
\label{sec:intro}
Much of the recent progress in large language model reasoning has been driven by online reinforcement learning (RL) methods such as PPO and GRPO during post-training. These methods instil self-correction, long chain-of-thought, and search-like behaviour into strong base models~\citep{shao2024deepseekmath, openai2024openaio1card, glm5team2026glm5vibecodingagentic}. Smaller and weaker base models typically inherit these behaviours through \emph{distillation} from the resulting teachers~\citep{yang2025qwen3technicalreport}.

Rollout generation during the RL phase dominates post-training compute: \citet{morrison2026hiddencostthinkingenergy} report that rollout generation accounts for 87\% of the energy used in OLMo 3 32B Think-specific post-training.
Ordinarily, despite taking most of the post-training compute, these rollouts (\Dpool{}) are discarded. Instead, canonical distillation pipelines spend additional compute on a separate generation pass by taking the final teacher model, generating new solutions for a prompt set, and filtering for correct responses via rejection sampling (\Dsynth{}). Students are then trained with supervised fine-tuning (SFT) to imitate teacher outputs~\citep{muennighoff2025s1simpletesttimescaling, guha2025openthoughtsdatarecipesreasoning, bespoke_stratos,deepseek2025r1}.

\begin{figure}[t]
  \centering  \includegraphics[width=0.72\linewidth]{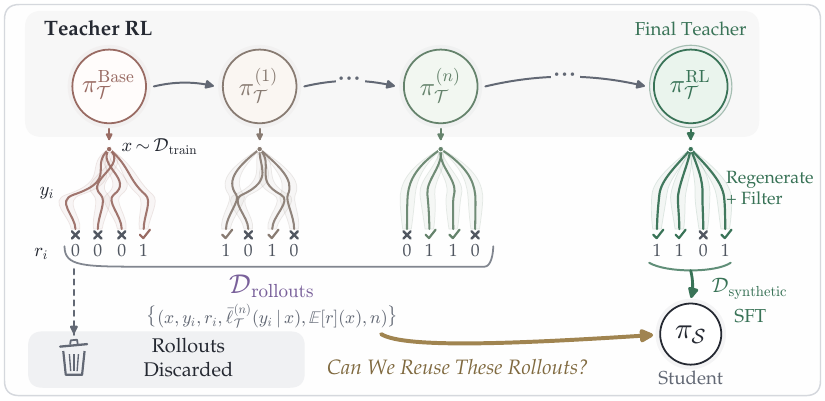}\par
  \vspace{5pt}
  \includegraphics[width=0.72\linewidth]{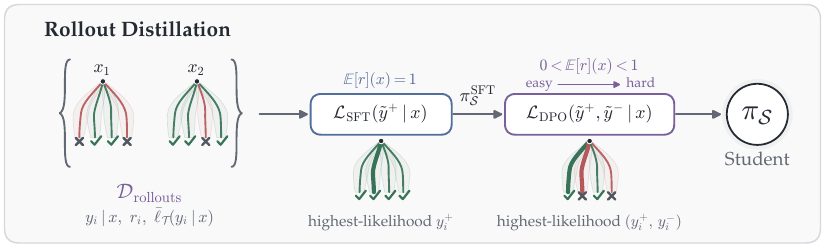}\par
  \caption{
    \textbf{Reusing teacher RL rollouts for student distillation.}
    \textbf{Top:} Canonical distillation discards the rollouts saved at every teacher RL step $n$ (completions $y_i$, rewards $r_i$, log-likelihoods $\bar{\ell}_{\mathcal{T}}^{(n)}$) and pays for a separate generation pass from the teacher (\Dsynth{}). \textbf{Bottom:} Our method curates the saved rollouts using the pass rate $\mathbb{E}[r](x)$: we first run SFT on the all-correct prompts ($\mathbb{E}[r](x)=1$), then DPO on the mixed-success prompts ($0<\mathbb{E}[r](x)<1$), ordered from easy to hard. Within each prompt, we rank the responses by their saved teacher log-likelihood $\bar{\ell}_{\mathcal{T}}$ and use the top correct $\tilde{y}^{+}$ and incorrect $\tilde{y}^{-}$.
  }
  \label{fig:rolloutdistillation}
  \label{fig:method-curation}
  \vspace{-1.0em}
\end{figure}

{Intuitively, this process seems wasteful:} %
teacher RL spans the same diverse, longer-horizon tasks such as maths, software engineering, and tool use that distillation targets~\citep{nemo_rl_ultra_recipe,nvidia_nemotron_3_ultra_2026}. Reusing these rollouts (Figure~\ref{fig:rolloutdistillation}) should remove the need for regeneration while exposing rich information like preference pairs and rewards already computed during teacher RL.

However, effectively reusing these data is non-trivial. Teacher rollouts are generated from evolving policies where potentially only the later steps have the reasoning behaviours that we want to instil in the student, and the teacher completions can be arbitrarily off-policy for the student~\citep{tang2024understandingperformancegaponline, tajwar2024preference, miao2024aligning, Shenfeld2025RLsRW, chu2025sft}. Consistent with previous work, we observe that naively using teacher rollouts for SFT underperforms canonical distillation. Similarly, we find that using the pairwise correct--incorrect information with Direct Preference Optimisation~\citep{rafailov2024dpo} is high variance, improving strong students but collapsing for weaker ones (Figure~\ref{fig:headline}).

In this paper we ask: \textit{Given access to a teacher's RL rollouts, how can we use them effectively for student distillation?} To answer this question, we treat rollout reuse as a data selection problem over the saved teacher RL rollouts using two pieces of metadata already recorded during GRPO: prompt pass rate and teacher log-likelihoods.

Inspired by the success of curriculum learning~\citep{gunasekar2023phi, abdin2024phi4technicalreport, mukherjee2023orcaprogressivelearningcomplex, wen-etal-2025-light}, we use the prompt pass rate both to split the archive into an SFT warm start on the easiest (all-correct) questions followed by DPO on mixed-success questions, and to order prompts from easy to hard. For each prompt we have multiple completions; we keep the responses with the highest cached teacher log-likelihood. This selection is based on the observation that student preferences agree with the teacher preferences cached during rollouts---the completions with the highest likelihood under the base student are also likely under the teacher, across all five students (Figure~\ref{fig:topk-correct-vs-incorrect}). Cached teacher likelihood is therefore a {student-agnostic} proxy for student-compatible completions~\citep{zhang2026the}, requiring no scoring pass over the archive for each new student.\looseness=-1

We compare our curation recipe against the canonical distillation pipeline on Python coding and maths.
Across students from various model families, our SFT$\to$DPO recipe on average matches canonical distillation, without any additional teacher generation or student scoring passes. In our setting, this avoids generating 58.3M maths tokens and 17.6M coding tokens, and such generation passes can be large at industrial scale: DeepSeek-R1, for example, reports about 600k reasoning samples kept after rejection sampling from an RL checkpoint~\citep{deepseek2025r1}. In addition, we study the failure of naive rollout usage, compare our selection rules with other curation baselines, and analyse when an SFT warm start is required before preference learning.

More broadly, we hope that these findings encourage the community to treat rollouts as a resource instead of discarding them. We ask practitioners to save and share their rollouts, so that the compute spent on RL for large models can also train smaller models for the ``compute-poor''.\footnote{Our code and saved rollouts are available at \url{https://github.com/vashistht/rollout_reuse}.}

\section{Related Work}
\label{sec:related}

\textbf{Reasoning distillation.}
Distillation by generating synthetic data from {strong} teachers underlies recent open reasoning models~\citep{abdin2024phi4technicalreport, yang2025qwen3technicalreport, deepseek2025r1, openthoughts-agent}; we keep this setup but ask if we can replace the separate generation pass with rollouts saved during teacher RL.

\textbf{Learning negative signals in RL rollouts.}
\citet{xu2025harnessingnegativesignalsreinforcement} also observe that negative traces can be useful and study how to better use traces from {the Open-R1} dataset, where they suggest asymmetric weighting. Our question is orthogonal: we ask how to \emph{select} which rollouts to imitate or contrast, using only metadata that is free under GRPO. {While their baseline is naive SFT on rollouts, we compare against the canonical separate-inference distillation pipeline (\Dsynth{}), treating naive SFT on rollouts as our lower bound.}

\textbf{Student-aligned data selection.} {\citet{zhang2026the}, in their work GRAPE,} show that aligning supervision with the student's own distribution improves SFT. They distil on the responses that the student finds most likely (Section~\ref{sec:proxy}), while our proxy uses scores recorded during teacher training. We compare the cost and performance of GRAPE and our proxy in Section~\ref{sec:selector-results}.

\section{Preliminaries}
\label{sec:background}

\paragraph{Teacher Training Using GRPO.} \label{sec:bg-teacher} When training a teacher policy $\pi_{\mathcal{T}}^\text{Base}$ with Group Relative Policy
Optimisation (GRPO)~\citep{shao2024deepseekmath}, we
sample a group $\{y_1,\ldots,y_G\}$ for a given prompt $x$, assign each completion a
reward $r_i\in\{0,1\}$ (for verifiable tasks), and compute the group-relative
advantage. The final teacher is referred to as $\pi_{\mathcal{T}}^\text{RL}$. For a generator policy $\pi_{\theta_{old}}$ and training policy $\pi_{\theta}$, the length-normalised, non-KL-regularised GRPO update is given by:
\begin{equation}
\begin{aligned}
\mathcal{J}_{GRPO}(\pi_{\theta})
= \mathbb{E}_{\hlx{x} \sim \Drl,\,
\{\hly{y_i}\}_{i=1}^G \sim \pi_{\theta_{old}}(\cdot|x)}&
 \frac{1}{G} \sum_{i=1}^G \frac{1}{|y_i|}
\sum_{t=1}^{|y_i|}
\Bigl\{ \min \Bigl[
\frac{\pi_{\theta}(y_{i,t}|x, y_{i,<t})}
{\hlpi{\pi_{\theta_{old}}(y_{i,t}|x, y_{i,<t})}}
\hlmu{\hat{A}_{i}}, \\
& \qquad \text{clip} \left(
\frac{\pi_{\theta}(y_{i,t}|x, y_{i,<t})}
{\pi_{\theta_{old}}(y_{i,t}|x, y_{i,<t})},
1 - \epsilon, 1 + \epsilon
\right) \hat{A}_{i}
\Bigr] \Bigr\}
\end{aligned}
\label{eq:Lgrpo}
\end{equation}
where $\hat{A}_{i}=\bigl(\hlr{r_i}-\mu_G(x)\bigr)/\sigma_G(x)$, with $r_i=R(x,y_i)$ the verifier reward and $\mu_G(x),\sigma_G(x)$ the mean and standard deviation of $\{r_j\}_{j=1}^G$. We pool $\mu_G(x)$ over steps into the pass rate $\hlmu{\mathbb{E}[r](x)}$.

In addition, for every $y_i$, we store its mean log-likelihood under the generator policy:
\begin{equation}
\bar{\ell}^{(n)}_{\mathcal{T}}(y_i\mid x)
=\frac{1}{|y_i|}\sum_{t=1}^{|y_i|}
\log \pi^{(n)}_{\mathcal{T}}(y_{i,t}\mid x,y_{i,<t}).
\label{eq:teacher-mean-logp}
\end{equation}

\paragraph{Creation of \Dpool{}.}
The highlighted quantities are computed for every GRPO update anyway. We use these by-products to construct \Dpool{}, a dataset of prompts ($x$), their completions ($y_i$) with corresponding rewards ($r_i$), pass rates ($\mathbb{E}[r](x)$), and mean log-likelihoods (Eq.~\eqref{eq:teacher-mean-logp}). Together, these give the rollout dataset, where $n$ runs over the RL steps up to the final teacher:
\begin{equation}
\Dpool=\{(\hlx{x},\hly{y_i},\hlr{r_i},\hlpi{\bar{\ell}^{(n)}_{\mathcal{T}}(y_i\mid x)},
\hlmu{\mathbb{E}[r](x)},n)\}.
\label{eq:defn-rollout}
\end{equation}

We note that for a GRPO run of $N$ training steps with batch size $B$, the total number of completions in \Dpool{} is $|\Dpool{}| = N \times B \times G$. At production scale this archive is large: the public Nemotron 3 Ultra recipe uses a global batch of 8,192 (512 prompts $\times$ 16 generations) for 178 steps, producing about 1.46M trajectories~\citep{nemo_rl_ultra_recipe}.%

\paragraph{Distillation Objectives.}

\label{sec:bg-objectives}
Canonical distillation samples from the final teacher ($y^+ \sim \pi_{\mathcal{T}}^\text{RL}$) and trains a
student policy $\pi_{\mathcal{S}}$ by SFT on correct completions,
\begin{equation}
\mathcal{L}_{\mathrm{SFT}}(\theta)=
-\mathbb{E}_{(x,y^+)\sim \mathcal{D}_{\mathrm{SFT}}}
\log \pi_{\mathcal{S}}(y^+\mid x).
\label{eq:sft}
\end{equation}

For a prompt $x$, let $P(x)=\{i:r_i=1\}$ and $N(x)=\{i:r_i=0\}$ denote its correct and incorrect completion indices. To construct data for SFT from \Dpool, we can filter for correct completions.
In addition to SFT, we can leverage the preference pairs from the rollouts for contrastive learning. Every mixed-success prompt from \Dpool{} supplies such pairs ($y^+$ from $P(x)$, $y^-$ from $N(x)$).
These pairs can be used to train a contrastive objective such as DPO~\citep{rafailov2024dpo}. Given a reference student policy
$\pi_{\mathcal{S}_\mathrm{ref}}$ and a pair $(y^+,y^-)$, DPO optimises
\begin{equation}
\begin{aligned}
\mathcal{L}_{\mathrm{DPO}}(\theta)
&=-\log\sigma\!\left(\beta\left[
\log\frac{\pi_{\mathcal{S}}(y^+\mid x)}{\pi_{\mathcal{S}_\mathrm{ref}}(y^+\mid x)}-\log\frac{\pi_{\mathcal{S}}(y^-\mid x)}{\pi_{\mathcal{S}_\mathrm{ref}}(y^-\mid x)}\right]\right).
\end{aligned}
\label{eq:dpo}
\end{equation}
DPO uses the contrastive signal and learns to widen the margin between $y^+$ and $y^-$: Eq.~\eqref{eq:dpo} depends on a \emph{difference} of student/reference log-likelihood ratios. A question then is which positive completion to imitate from $P(x)$ and which of the $|P(x)|\times|N(x)|$ preference pairs to use for DPO.

\section{Motivation}
\label{sec:motivation}
After Section~\ref{sec:background}, you are hopefully convinced that \Dpool{} is a valuable resource, especially when coupled with information like likelihoods and average pass rates. Using it directly for distillation is, however, non-trivial. In this section we first discuss why, and then motivate our solutions.
\subsection{Difficulty in Using Rollout Data Naively}
\label{sec:why-naive-is-hard}
As seen in Eq.~\eqref{eq:sft}, during SFT the student learns to mimic the teacher's distribution. In canonical distillation, this target is the final teacher that regenerates the data:
$
y_{\mathrm{regeneration}} \sim\pi_{\mathcal T}^{\mathrm{RL}}(\cdot\mid x).
$
On the other hand, rollout reuse draws from the union of the responses saved throughout teacher training, $\Dpool=\bigcup_n\Dpool^{(n)}$ with $y^{(n)}\sim\pi_{\mathcal T}^{(n)}(\cdot\mid x)$ at step $n$.
Intuitively, the student is therefore asked to imitate a mixture of policies: earlier policies, even when correct, may not yet exhibit behaviours acquired later in RL, and the teacher changes substantially over training (Figure~\ref{fig:teacher-train-dynamics}; Appendix~\ref{app:temporal-provenance}). This motivates selecting which correct rollouts the student should imitate.

\subsection{Off-Policy Methods (SFT/DPO) and the Need for Data Selection}
\label{sec:offpolicy-data}
Both SFT (Eq.~\eqref{eq:sft}) and DPO (Eq.~\eqref{eq:dpo}) are off-policy: the policy that generated the data is not the one being trained. During SFT, the student learns to imitate $y^+$, but the distribution of $y^+$ can be arbitrarily far from the student's own. Previous work has shown that imitating a distribution far from the student's pre-training distribution can degrade generalisation and induce forgetting~\citep{Shenfeld2025RLsRW, chu2025sft}.

\paragraph{DPO dependence on student support.}
DPO has the same issue, as it learns to separate $y^+$ from $y^-$ when either can be far off-policy for the student, with mistakes the student itself may never make. As seen in Eq.~\eqref{eq:dpo}, at initialisation ($\pi_{\mathcal{S}} = \pi_{\mathcal{S}_\mathrm{ref}}$),
the implicit reward gap starts at zero regardless of absolute likelihood; \citet{yang2025mitigatinghallucinationslargevisionlanguage} show that when both completions lie far outside the base student's support, early
gradient steps push the policy into a region where the log-ratio
$\beta\log\frac{\pi_{\mathcal{S}}(y|x)}{\pi_{\mathcal{S}_\mathrm{ref}}(y|x)}$
grows large in magnitude and the sigmoid can saturate. This can
stall preference learning before the model learns to meaningfully separate the two traces~\citep{yang2025mitigatinghallucinationslargevisionlanguage}.
Thus, preference learning might be difficult when completions lie far outside the base student's support. This motivates an SFT warm start on the easy questions before DPO, so that the chosen responses are already likely under the student when preference learning begins. This leads to our first hypothesis ($\mathbb{H}_1$), tested in Section~\ref{sec:q3}:

\begin{stylehypothesis}{$\pmb{\mathbb{H}}_{\mathbf{1}}$ (Student Support for Preference Learning).} DPO on the rollouts may fail for students that assign low likelihood to the chosen responses; an SFT warm start that makes these responses likely before DPO can potentially help recover these cases.
\end{stylehypothesis}

\subsection{Student-Compatible Traces and Approximating Them with Cached Teacher Likelihoods}\label{sec:proxy}
To tackle some of these idiosyncrasies,
GRAPE~\citep{zhang2026the} gathers responses from various sources and keeps the one closest to the base student's distribution. Given correct completions $y_i$, $i\in P(x)$, each of length $T_i=|y_i|$, and base student $\mathcal{S}_0$, GRAPE chooses the highest probability completion:
\begin{equation}
i^\star_{\mathcal{S}}(x)=\operatorname*{arg\,max}_{i\in P(x)}\frac{1}{T_i}\log \mathbb{P}_{\mathcal{S}_0}(y_{i,1:T_i}\mid x)
=\operatorname*{arg\,max}_{i\in P(x)}\bar{\ell}_{\mathcal{S}_0}(y_i\mid x),
\qquad y^\star_{\mathcal{S}}(x)=y_{i^\star_{\mathcal{S}}(x)},
\label{eq:grape}
\end{equation}
where $\bar{\ell}_{\mathcal{S}_0}$ is the mean completion-token log-likelihood of Eq.~\eqref{eq:teacher-mean-logp} under the base student.

As can be seen from Eq.~\eqref{eq:grape}, creating the GRAPE dataset from rollouts would require a pass over the full rollout pool for \emph{each student}. Therefore, ideally, we wish to have a student-agnostic way to select completions.

\begin{figure}[!t]
    \centering
    \includegraphics[width=\linewidth]{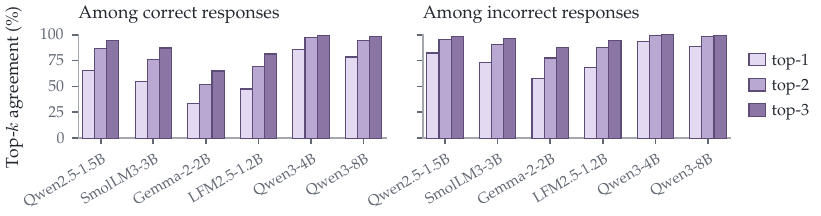}
    \vspace{-0.5em}
    \caption{%
    \textbf{Agreement between cached teacher and base-student preferences ($\mathcal{T}_1$).}
    Fraction of prompts where the base student's preferred completion ($\arg\max\bar{\ell}_{\mathcal{S}_0}$) lies in the teacher's top-$k$ by cached mean log-likelihood $\bar{\ell}^{(n)}_{\mathcal{T}}$ ($k\in\{1,2,3\}$), shown separately for correct (left) and incorrect (right) responses. Across all student families, completions with the highest likelihood under the teacher policy are also highly likely under the student, motivating our proxy in Section~\ref{sec:method}.
}
\label{fig:topk-correct-vs-incorrect}\label{fig:topk-agreement}
    \vspace{-0.8em}
\end{figure}

\paragraph{Approximating student preferences with cached teacher likelihoods.}
We use the cached normalised log-likelihoods
(Eq.~\eqref{eq:teacher-mean-logp}) that we are already saving in \Dpool{} as a student-agnostic proxy.
The intuition is that teacher and student rank completions similarly (if one is preferred under the teacher policy, it is also more likely under the student policy than the other completions):
\begin{equation}
\operatorname{rank}_{\mathcal{S}}(y_i;x)
\;\approx\;
\operatorname{rank}_{\mathcal{T}}(y_i;x),
\label{eq:ranking-relationship}
\end{equation}
where the student and teacher rankings are induced by
$\bar{\ell}_{\mathcal{S}_0}(y_i\mid x)$ and
$\bar{\ell}^{(n)}_{\mathcal{T}}(y_i\mid x)$, respectively. To test whether this holds across students, we score every saved completion with each base student and, within each GRPO group, report \emph{top-$k$ agreement}: how often the student's preferred completion lies among the teacher's top $k$ (top-1 agreement means that both select the same completion for a given prompt). Figure~\ref{fig:topk-correct-vs-incorrect} shows this on $\mathcal{T}_1$ for our five base students and Qwen3-8B-Base, from four model families: as expected, agreement is highest for the in-family Qwen3 students and lowest for Gemma-2-2B. This leads to our second hypothesis ($\mathbb{H}_2$):

\begin{stylehypothesis}{$\pmb{\mathbb{H}}_{\mathbf{2}}$ (Likelihood-Rank Agreement).}
Teacher and student preferences over the set of completions of a prompt are highly correlated; therefore, the teacher-top1 rollout is a better training target than a random correct rollout.
\end{stylehypothesis}

We compare the downstream performance of teacher-top1 and GRAPE (student-top1) in maths and coding in Section~\ref{sec:selector-results}.

\section{Method}
\label{sec:method}
Building on the observations in Section~\ref{sec:motivation}, we treat rollout reuse as a data selection problem using metadata already recorded during GRPO, namely the pooled prompt pass rate
and the cached log-likelihood%
. This framing raises the following three questions:

\textbf{\textit{$Q_1$: Which questions should we reuse, and in what order?}}
We use $\mathbb{E}[r](x)$ as a proxy for prompt difficulty and partition \Dpool{} into consistently solved prompts ($\mathbb{E}[r](x)=1$) for an SFT warm start and mixed-success prompts ($0<\mathbb{E}[r](x)<1$) for DPO. All-correct prompts provide a natural candidate for SFT, while mixed-success prompts provide the correct--incorrect pairs for DPO. Following curriculum learning, we order the mixed-success prompts from easy to hard by decreasing $\mathbb{E}[r](x)$.

\textbf{\textit{$Q_2$: Given multiple completions for a question, how do we select which ones to train on?}}
Rather than scoring all rollouts under each new student as in GRAPE, we exploit the rank agreement between teacher and student preferences (Eq.~\eqref{eq:ranking-relationship}) and select the highest-likelihood correct and incorrect rollouts under the generating teacher policy (\emph{teacher-top1}). Maximising $\bar{\ell}^{(n)}_{\mathcal{T}}$ selects student-compatible positives and plausible, in-support negatives (Figure~\ref{fig:topk-correct-vs-incorrect}, right).

\textbf{\textit{$Q_3$: Rollouts have both positive and negative completions; if and when does preference learning on these help over SFT?}}
DPO on its own works well for many students, but it fails for the students that assign the lowest likelihood to the chosen responses. Because DPO can stall or collapse when chosen responses lie far outside the base student's support, we first warm-start the student with SFT on the easiest prompts, and then initialise DPO from this checkpoint on the curriculum-ordered preference pairs.
\begin{algorithm}[h!]
\caption{Two-stage rollout distillation (\DtwoStage{}) from \Dpool{} for a base student $\pi_{\mathcal{S}_0}$}
\label{alg:recipe}
\begin{algorithmic}[1]
\small
\Require rollout dataset \Dpool{} (Eq.~\eqref{eq:defn-rollout}) of prompts $\hlx{x}$, completions $\hly{y_i}$, rewards $\hlr{r_i}$, mean log-likelihoods $\hlpi{\bar{\ell}^{(n)}_{\mathcal{T}}(y_i\mid x)}$ and pass rates $\hlmu{\mathbb{E}[r](x)}$; base student $\pi_{\mathcal{S}_0}$.
\Statex $P(x)$, $N(x)$ as in Section~\ref{sec:background}.
\State \textbf{Split \Dpool{} by pass rate} into all-correct and mixed-success prompts: \Comment{$Q_1$}
\Statex \hspace{\algorithmicindent}$\mathcal{X}^{\mathrm{allp}}\gets\{x:\hlmu{\mathbb{E}[r](x)}=1\}$, \quad $\mathcal{X}^{\mathrm{mix}}\gets\{x:0<\hlmu{\mathbb{E}[r](x)}<1\}$
\State \textbf{Use the highest-likelihood $y^+$ and $y^-$} of each prompt: \Comment{$Q_2$}
\Statex \hspace{\algorithmicindent}$\tilde y^{+}_x\gets\operatorname*{arg\,max}_{y_i:\,i\in P(x)}\hlpi{\bar{\ell}^{(n)}_{\mathcal{T}}(y_i\mid x)}$, \quad $\tilde y^{-}_x\gets\operatorname*{arg\,max}_{y_i:\,i\in N(x)}\hlpi{\bar{\ell}^{(n)}_{\mathcal{T}}(y_i\mid x)}$
\State \textbf{Build the SFT data} from the all-correct prompts: \ $\DsftAllp\gets\{(x,\tilde y^{+}_x):x\in\mathcal{X}^{\mathrm{allp}}\}$
\State \textbf{Build the DPO data} from the mixed prompts, ordered from easy to hard (decreasing $\hlmu{\mathbb{E}[r](x)}$): \Comment{$Q_1$}
\Statex \hspace{\algorithmicindent}$\DdpoCurr\gets\{(x,\tilde y^{+}_x,\tilde y^{-}_x):x\in\mathcal{X}^{\mathrm{mix}}\}$
\State \textbf{Run SFT} on the base student: \ $\pi^{\mathrm{SFT}}_{\mathcal{S}}\gets\mathrm{SFT}(\pi_{\mathcal{S}_0};\DsftAllp)$ \Comment{$Q_3$}
\State \textbf{Run DPO} from the SFT checkpoint: \ $\pi_{\mathcal{S}}\gets\mathrm{DPO}(\pi^{\mathrm{SFT}}_{\mathcal{S}};\DdpoCurr)$ \Comment{$Q_3$}
\end{algorithmic}
\end{algorithm}

The answers to these questions lead us to our method (Algorithm~\ref{alg:recipe}), which uses metadata cached during teacher RL. %
Notably, this method does not require any new generations from the teacher or any student-specific forward passes for data selection and creation.

\FloatBarrier %
\section{Experimental Setup}
\label{sec:exp}
\textbf{Setting.}
We study rollout reuse in two domains with verifiable rewards, mathematical reasoning and Python coding. As described in Figure~\ref{fig:rolloutdistillation}, the setup includes first training a teacher model with GRPO, then distilling into students either from the final teacher or by reusing the saved rollouts, and evaluating the distilled students on a set of benchmarks.

\textbf{Training Datasets for Teacher RL.} \label{sec:exp-datasets} In maths, we use \textbf{Math12K}~\citep{hendrycks2021measuringmathematicalproblemsolving}, the 12,000 MATH training problems in the split of \citet{lightman2023letsverifystepstep} (holding out MATH500), and \textbf{DeepScaler}~\citep{deepscaler2025}, a broader collection of competition maths problems, from which we use a random 20,000-problem subset. For coding, we use \textbf{AceCode-87K}~\citep{zeng2025acecoderacingcoderrl}: 20,480 Python training tasks and a disjoint 256-task validation split after deduplication and test-set decontamination (rewarded only if all tests pass).

\textbf{Teacher RL.}
\label{sec:exp-settings}
We train all teachers with GRPO in verl~\citep{sheng2024hybridflow}, with $G{=}16$ rollouts per prompt and binary verifier rewards. We instantiate two maths teachers that differ in the relationship between the teacher's RL prompt distribution and the student's evaluation distribution:
\begin{itemize}[leftmargin=1.5em, itemsep=1pt, topsep=2pt]
    \item $\mathcal{T}_1$ --- \textbf{in-distribution} ($\Drl \stackrel{d}{=} \mathcal{D}_\mathrm{eval}$). The teacher (Qwen3-4B-Base) is trained with full-parameter GRPO on the Math12K training split, and the student is evaluated on MATH500. The saved rollouts therefore come from the same distribution as the test prompts ($\therefore$ in-dist.).
    \item $\mathcal{T}_2$ --- \textbf{cross-distribution} ($\Drl \stackrel{d}{\neq} \mathcal{D}_\mathrm{eval}$). The teacher (Qwen3-8B-Base) is trained with GRPO using LoRA~\citep{hu2022lora} on DeepScaler-20K~\citep{deepscaler2025}, and the student is again evaluated on MATH500. The saved rollouts therefore come from a related but different distribution than the test prompts ($\therefore$ out-dist.), making this the harder setting for rollout distillation.
\end{itemize}
In coding, we fully fine-tune Qwen3-4B-Base on AceCode and evaluate students on HumanEval+ and MBPP+. We choose the final teachers on validation performance: step 234 for $\mathcal{T}_1$ and step 120 for $\mathcal{T}_2$ and coding (Appendix~\ref{app:training}).

\begin{table}[h]
\centering\small
\setlength{\tabcolsep}{5pt}
\begin{tabular}{@{}lllll@{}}
\toprule
Dataset & Stage & Prompts & Order & Completion selector \\
\midrule
\DsftRand{} & SFT & $\mathbb{E}[r](x)\in(0,1]$ & shuffled & random \\
$\mathcal{D}_{\text{SFT}}^{\text{shuffled, top1}}$ & SFT & $\mathbb{E}[r](x)\in(0,1]$ & shuffled & teacher-top1 \\
GRAPE & SFT & $\mathbb{E}[r](x)\in(0,1]$ & shuffled & student-top1 \\
\DsftCurr{} & SFT & $\mathbb{E}[r](x)\in(0,1]$ & easy $\to$ hard & teacher-top1 \\
\DsftAllp{} & SFT & $\mathbb{E}[r](x)=1$ & --- & teacher-top1 \\
\midrule
\DdpoRand{} & DPO & $\mathbb{E}[r](x)\in(0,1)$ & shuffled & random \\
$\mathcal{D}_{\text{DPO}}^{\text{shuffled, top1}}$ & DPO & $\mathbb{E}[r](x)\in(0,1)$ & shuffled & teacher-top1 \\
\DdpoCurr{} & DPO & $\mathbb{E}[r](x)\in(0,1)$ & easy $\to$ hard & teacher-top1 \\
\bottomrule
\end{tabular}
\vspace{4pt}
\caption{\textbf{Descriptions of the rollout datasets studied.} Each dataset keeps one correct completion per prompt for SFT or one correct--incorrect pair for DPO. Following $Q_1$--$Q_2$, we vary \emph{which questions} to use (by pass rate $\mathbb{E}[r](x)$), \emph{what order} to present them in (shuffled vs.\ easy $\to$ hard), and \emph{which completions} to select: random, cached teacher log-likelihood (teacher-top1), or base-student log-likelihood (student-top1; GRAPE~\citep{zhang2026the}).}
\label{tab:variants}
\end{table}

\textbf{Student Models.}
\label{sec:exp-models}
We distil our teachers into five base students from several families and architectures: Qwen2.5-1.5B~\citep{qwen2.5} and Qwen3-4B-Base~\citep{yang2025qwen3technicalreport} from the Qwen family, SmolLM3-3B-Base~\citep{bakouch2025smollm3} and Gemma-2-2B~\citep{gemmateam2024gemma2} as non-Qwen transformers, and LFM2.5-1.2B-Base~\citep{liquidai2025lfm2} as a hybrid student. In coding, we use the same students except Gemma-2-2B (given its 8K context limit).

\textbf{Baselines and Rollout Datasets.}
\label{sec:variants}
Our main baseline is canonical distillation, \Dsynth{}: we sample eight responses per training prompt from the final teacher and train on the correct ones (one random correct response per prompt in coding), regenerating on Math12K for both maths teachers and on AceCode in coding. From \Dpool{}, we compare our recipe with naive random baselines (\DsftRand{}, \DdpoRand{}), GRAPE, and other variants (the datasets are detailed in Table~\ref{tab:variants}).

\textbf{Training Students.}
\label{sec:exp-training}
We use TRL~\citep{vonwerra2020trl} to train SFT with AdamW and DPO with $\beta=0.1$. For \Dsynth{}, we sweep the learning rates and the scheduler ($10^{-5}$ or $5\times10^{-5}$, constant or cosine) and keep the best run for each student; rollout SFT uses $10^{-5}$ with a cosine schedule. DPO runs use $5\times10^{-6}$ with a constant schedule, chosen on \DdpoRand{} (Appendix~\ref{app:training}).

\textbf{Evaluations.}
In maths, we evaluate the distilled students on MATH500 from the OpenAI split~\citep{lightman2023letsverifystepstep, hendrycks2021measuringmathematicalproblemsolving}, GSM8K~\citep{cobbe2021training} (1,319 grade-school word problems), and AIME 2024, AIME 2025 and AMC 2023 \citep{maa_competitions}, graded with Math-Verify~\citep{kydlicek2024mathverify}. In coding, we evaluate on HumanEval+ (164 problems) and MBPP+ (378 problems)~\citep{liu2023evalplus}, which add tests to HumanEval~\citep{chen2021evaluating} and MBPP~\citep{austin2021program}. We report the average accuracy over 8 samples at temperature $0.7$ (avg@8). Checkpoints are selected on MATH500 in maths and on the AceCode validation split in coding. In addition, we report one standard deviation as error bars and 95\% confidence intervals from a paired bootstrap over test problems (Appendix~\ref{app:full-results}).

\section[Results]{Results\label{sec:results}}

\subsection[Closing the Gap to Canonical Distillation]{Closing the Gap to Canonical Distillation\label{sec:results-naive}\label{sec:q1}\label{sec:main-performance}\label{sec:results-headline}\label{sec:coding-results}}

\begin{figure}[!ht]
    \centering
    \includegraphics[width=.9\linewidth]{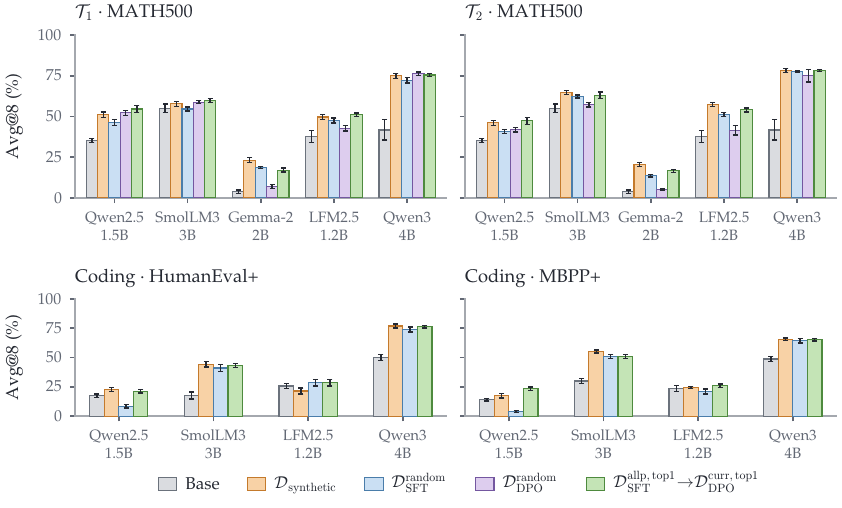}
    \caption{\textbf{Comparison of canonical distillation and rollout reuse.} Avg@8 on MATH500 for $\mathcal{T}_1$ (Math12K rollouts) and $\mathcal{T}_2$ (DeepScaler rollouts) in the top row, and on HumanEval+ and MBPP+ in coding (bottom row); base students are shown in grey. \Dsynth{} is generated on Math12K for both maths teachers. Naive rollout reuse falls short of canonical distillation, and our two-stage recipe closes most of the gap.}
    \label{fig:headline}\label{fig:accuracy-vs-synth}\label{fig:headline-code}\label{fig:coding-student-accuracy}
\end{figure}

\textbf{Naive rollout reuse underperforms canonical distillation.} Figure~\ref{fig:headline} compares distillation from the saved rollouts against canonical distillation (Table~\ref{tab:main-results}). Consistent with our motivation, naively training SFT on a random correct rollout per prompt (\DsftRand{}) underperforms \Dsynth{} across all ten maths teacher--student settings and in seven of eight coding comparisons ($-1.95$ points in maths and $-4.44$ in coding, averaged across students and benchmarks).
Training DPO on random preference pairs (\DdpoRand{}) is unstable: while it improves over \Dsynth{} for three students on $\mathcal{T}_1$, it collapses to near-base accuracy on Gemma-2-2B and underperforms \Dsynth{} for every student on $\mathcal{T}_2$. Simply having a large rollout pool is therefore not enough; how we select from \Dpool{} matters.

\textbf{Our recipe closes most of the gap.} Our two-stage recipe (\DtwoStage{}) improves over both naive rollout baselines for nearly every student. Averaged across students and benchmarks, it matches \Dsynth{} closely in maths ($-0.37$ points, compared with $-1.95$ for \DsftRand{}) and is on par with \Dsynth{} in coding ($+0.89$ points;
refer to Appendix~\ref{app:full-results} for CI). It also outperforms GRAPE-SFT in six of eight maths settings and six of eight coding cells (Table~\ref{tab:main-results}), without requiring any new teacher generation (Table~\ref{tab:generation-cost}) or student forward passes over the rollouts.

\begin{table}[h]
\centering
\small
\setlength{\tabcolsep}{6pt}
\begin{tabular}{@{}lrr@{}}
\toprule
Completions / response tokens & Maths ($\mathcal{T}_1$, step 234) & Coding (step 120) \\
\midrule
Saved during teacher RL, reused by our recipe & 239,592 / 167.7M & 61,440 / 26.7M \\
New teacher generation for canonical distillation & 96,000 / 58.3M & 163,840 / 17.6M \\
New teacher generation for our recipe & 0 / 0 & 0 / 0 \\
\bottomrule
\end{tabular}
\vspace{4pt}
\caption{\textbf{Teacher generation cost of canonical distillation versus our recipe.} Total completions and response tokens (in millions, M) for the Qwen3-4B-Base maths ($\mathcal{T}_1$) and coding teachers. Canonical distillation samples eight new responses per prompt from the final teacher, whereas our recipe reuses rollouts already generated during teacher RL, saving millions of generated tokens.}
\label{tab:generation-cost}
\end{table}

\textbf{Cross-dataset transfer.} Unlike $\mathcal{T}_1$ (Math12K rollouts), $\mathcal{T}_2$ trains on DeepScaler, while its \Dsynth{} baseline regenerates solutions from the final 8B teacher on the same Math12K training prompts used by the $\mathcal{T}_1$ baseline.
Even under this harder test, our recipe is $1.5$ points below \Dsynth{} on average on MATH500, whereas naive SFT (\DsftRand{}) is $4.3$ points below. While the average gap to \Dsynth{} is slightly larger than in the in-distribution setting ($\mathcal{T}_1$), rollouts from a related prompt pool transfer effectively.

\subsection[The Need for Data Selection]{The Need for Data Selection\label{sec:q2}\label{sec:curation-results}\label{sec:selector-results}\label{sec:decoupling}\label{sec:generation-cost}\label{sec:q-prompts}\label{sec:quality-quantity}}

\begin{figure}[!t]
    \centering
    \includegraphics[width=.9\linewidth]{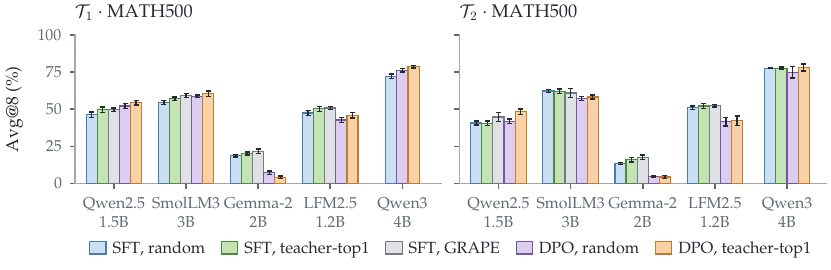}%
    \caption{\textbf{Comparison of different completion selection methods.} MATH500 avg@8 for $\mathcal{T}_1$ (Math12K rollouts, left) and $\mathcal{T}_2$ (DeepScaler rollouts, right), for SFT and DPO using random, teacher-top1, and GRAPE (student-top1) selection on the same prompts in shuffled order (Table~\ref{tab:selection-order}). We see that teacher-top1 selection improves over random selection.}
    \label{fig:selection}
\end{figure}

\textbf{Teacher-top1 selection improves over random selection.} To test the effect of the selection rule, we train on the exact same prompts using either a random rollout or our proxy (the teacher-top1 rollout). Teacher-top1 improves both SFT and DPO on average under both maths teachers, and improves SFT in coding as well (Table~\ref{tab:acecode-multistudent}). This is consistent with $\mathbb{H}_2$: the rollout the teacher found most likely is also high-likelihood under the student, making it a better target for student training than a random completion.

\textbf{GRAPE shows gains at the cost of a student-scoring pass.} GRAPE needs to score every correct rollout in the large rollout archive under each base student. On the same prompts ($\mathbb{E}[r](x)>0$), GRAPE achieves higher accuracy than teacher-top1 in maths ($+1.05$ and $+1.23$ points on MATH500 for $\mathcal{T}_1$ and $\mathcal{T}_2$; Figure~\ref{fig:selection}), while in coding teacher-top1 is on par ($+0.43$ points; Appendix~\ref{app:selector-cohorts}). On the student with the lowest likelihood overlap, Gemma-2-2B (Figure~\ref{fig:topk-correct-vs-incorrect}), GRAPE-SFT also beats our two-stage recipe, by $4.7$ and $1.1$ points for $\mathcal{T}_1$ and $\mathcal{T}_2$.
This shows that GRAPE gives us gains at
the cost of extra student passes over the rollout pool, while ours is zero-pass and student-agnostic.
Both methods show that rollouts are worth saving, which is the broader point of our paper.

\textbf{Prompt order matters little.} Our curriculum datasets combine two choices: teacher-top1 selection and easy-to-hard ordering by pass rate. Holding the selected completions fixed, ordering prompts from easy to hard instead of shuffling them gives only a small average gain on MATH500 ($+0.2$ points in SFT and $+0.9$ in DPO; Table~\ref{tab:selection-order}), helping most in DPO on $\mathcal{T}_2$ and for the weaker students, where standalone DPO is most fragile. Since pass rates are cached, our recipe orders prompts at no extra cost. We see that the completion selection rules discussed above matter more than prompt order ($+1.5$ points in SFT and $+1.7$ in DPO).

\subsection[When Does Preference Learning Help?]{When Does Preference Learning Help?\label{sec:weak-dpo}\label{sec:support-diagnostic}\label{sec:q3}\label{sec:q-objective}}

\begin{figure}[!ht]
    \centering    \includegraphics[width=0.9\linewidth]{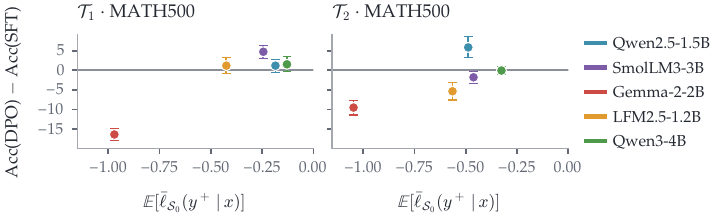}
    \vspace{-0.6em}
    \caption{\textbf{DPO versus SFT against base-student support.} Difference in MATH500 avg@8 between \DdpoCurr{} and \DsftCurr{}, which use the teacher-top1 correct responses, for $\mathcal{T}_1$ (left) and $\mathcal{T}_2$ (right), against \emph{base-student support} $\mathbb{E}[\bar{\ell}_{\mathcal{S}_0}(y^+\mid x)]$: the mean log-likelihood of the chosen DPO responses under the base student. DPO falls behind SFT for the students with the lowest base-student support (Gemma-2-2B, and LFM2.5-1.2B under $\mathcal{T}_2$); this trend is more apparent under $\mathcal{T}_1$.
    }
\label{fig:dpo-vs-sft}\label{fig:dpo-vs-sft-both}
    \vspace{-0.6em}
\end{figure}

\textbf{Preference learning helps most students, but fails when base-student support is low.} To test whether learning to contrast correct and incorrect responses (DPO) helps over imitating the correct ones (SFT), we compare \DdpoCurr{} with \DsftCurr{}. They both use teacher-top1 correct responses ordered from easy to hard (Figure~\ref{fig:dpo-vs-sft}). DPO matches or outperforms SFT for most students, but leaves Gemma-2-2B near its base accuracy, and on $\mathcal{T}_2$ it also falls behind SFT for LFM2.5-1.2B. Consistent with $\mathbb{H}_1$, these are the students whose base policies assign the lowest log-likelihood to the chosen responses. As discussed in Section~\ref{sec:offpolicy-data}, this motivates an SFT warm start to increase the student's support for correct responses before DPO.

\textbf{An SFT warm start helps the low-support students.} For the higher-support students (Qwen2.5-1.5B, SmolLM3-3B and Qwen3-4B), standalone DPO is already strong, and adding the warm start changes MATH500 by at most three points either way. For the low-support students, the warm start is useful: it lifts Gemma-2-2B from 4.3\% to 17.1\% on MATH500 under $\mathcal{T}_1$ and improves LFM2.5-1.2B under both teachers (Tables~\ref{tab:comprehensive-t1} and~\ref{tab:comprehensive-t2}).
This supports the second part of $\mathbb{H}_1$: an SFT warm start can recover the students for which standalone DPO fails.

\textbf{The two-stage recipe is reliable across students.} We see that neither stage works for all students on its own: standalone DPO fails on the low-support students, while the SFT warm start alone underperforms the full recipe for most students. Our two-stage recipe is therefore a reliable way to reuse rollouts: it stays within $1.3$ points of the better of the two stages in every maths setting and outperforms running both stages on random rollouts (\DsftRand{}$\to$\DdpoRand{}) in eight of ten maths settings on MATH500 (Appendix~\ref{app:full-math-results}).

\section{Discussion}
\label{sec:discussion}

We showed that the separate generation pass of canonical distillation can be replaced by curating saved teacher RL rollouts: selecting from \Dpool{} using the pass rates $\mathbb{E}[r](x)$ and log-likelihoods $\bar{\ell}^{(n)}_{\mathcal{T}}$ that GRPO already computes matches canonical distillation on average across maths and coding, whereas naive rollout reuse falls short. While spending extra compute to score the archive under each student (as in GRAPE) yields a small additional gain in maths, cached teacher likelihood provides a zero-cost, student-agnostic alternative. At frontier scale, where a single RL run spans millions of trajectories across diverse environments, the saved rollout pool can distil students without additional teacher inference.

\textbf{Limitations.}\label{sec:limitations} Our experiments study 4B and 8B teachers distilling into 1.2B--4B students; whether our curation rules transfer across much larger teacher--student capacity gaps remains an open question. We focus on mathematical reasoning and Python coding with binary verifier rewards; in domains with non-verifiable rewards, partitioning prompts into positive and negative traces is less well defined. Finally, while we report evaluation variance over 8 repeats, we run a single training seed per configuration.

\textbf{Future Directions.} Our study opens two natural directions for making fuller use of teacher training artefacts. First, we can test whether $\mathrm{SFT}\to\mathrm{DPO}$ rollout distillation provides a stronger initialisation than standard SFT distillation for subsequent on-policy distillation or RL. Second, teacher training also produces intermediate checkpoints $\pi_{\mathcal{T}}^{(1)}, \dots, \pi_{\mathcal{T}}^{(N)}$; using them and their stage-specific rollouts for progressive curriculum (on/off-policy) distillation is another promising direction.

We hope our findings encourage practitioners to save and share their RL rollouts along with information like rewards and likelihoods computed during RL---in short, don't discard your rollouts; reuse them for distillation.

\begin{ack}
This research was funded by the \href{https://ror.org/05xpvk416}{National Institute of Standards and Technology} under Federal Award ID Number 60NANB24D231 and \href{https://ror.org/05x2bcf33}{Carnegie Mellon University AI Measurement Science and Engineering Center (AIMSEC)}.
This material is also based upon work supported by the National Science Foundation \href{https://www.nsf.gov/awardsearch/showAward?AWD_ID=2326610}{Grant 2326610}. Any opinions, findings, and conclusions or recommendations expressed in this material are those of the author(s) and do not necessarily reflect the views of the National Science Foundation.
The authors thank the CMU FLAME Center and the CMU Babel Compute Cluster for computational support. We also thank Lintang Sutawika and Xiaochuan Li for insightful discussions, and Petros Karypis for feedback on the manuscript.
\end{ack}

\bibliographystyle{plainnat}
\bibliography{references}

\newpage
\appendix
\raggedbottom

\section{Experimental Details}
\label{app:training}

\subsection{Teacher Training}
\begin{figure}[htbp]
    \centering
    \includegraphics[width=0.95\linewidth]{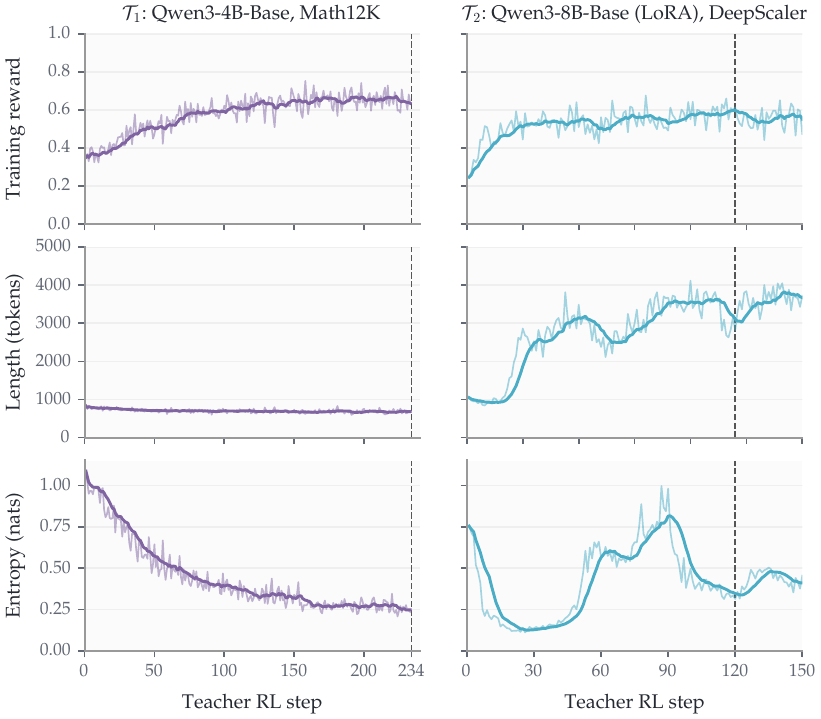}
    \caption{\textbf{Maths teacher training.} Training reward, response length and policy entropy of $\mathcal{T}_1$ (left) and $\mathcal{T}_2$ (right) during GRPO. Faint lines show each step and solid lines a ten-step moving average. Dashed lines mark the final teachers (steps 234 and 120), up to which we save the rollouts; $\mathcal{T}_2$ is shown up to step 150.}
    \label{fig:teacher-train-dynamics}
\end{figure}

We train all teachers with GRPO in verl~\citep{sheng2024hybridflow}, with $G{=}16$ rollouts per prompt, sampling temperature $1.0$ and no KL penalty. In maths, we consider two teacher settings, grading final answers with Math-Verify~\citep{kydlicek2024mathverify}:
\begin{itemize}[leftmargin=1.5em,itemsep=2pt,topsep=2pt]
    \item \textbf{$\mathcal{T}_1$} (train: Math12K; test: MATH500 and the other maths benchmarks). We fully fine-tune Qwen3-4B-Base for 234 steps on eight L40S GPUs, with learning rate $10^{-6}$, 64 prompts per step and a 3,072-token response cap. We see a plateau in MATH500 validation accuracy during teacher training, so we use the last step (234) as the final teacher.
    \item \textbf{$\mathcal{T}_2$} (train: DeepScaler-20K; test: the same benchmarks). We train Qwen3-8B-Base with LoRA ($r{=}128$, $\alpha{=}64$), learning rate $5\times10^{-5}$, 64 prompts per step, and an 8,192-token response cap on eight L40S GPUs (tensor parallelism 2) for 205 steps. We use step 120, its best MATH500 validation checkpoint, as the final teacher.
\end{itemize}
Figure~\ref{fig:teacher-train-dynamics} shows the training reward, response length and policy entropy of both teachers.

\begin{figure}[htbp]
\centering
\begin{minipage}[t]{0.49\linewidth}\centering\includegraphics[width=\linewidth]{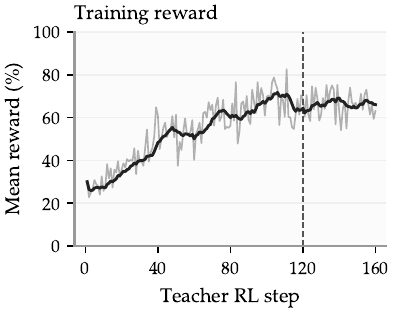}\end{minipage}\hfill
\begin{minipage}[t]{0.49\linewidth}\centering\includegraphics[width=\linewidth]{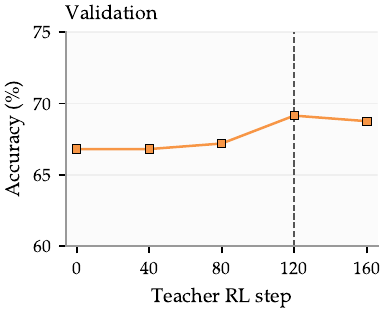}\end{minipage}
\caption{\textbf{Coding teacher training.} Left: mean training reward, per step (faint) and as a ten-step moving average (solid). Right: validation accuracy on the 256 held-out AceCode tasks, with greedy decoding, every 40 steps. The dashed line marks the final teacher at step 120, up to which we save the rollouts.}
\label{fig:ace-teacher-current}
\end{figure}

In coding (train: AceCode; test: HumanEval+ and MBPP+), we fully fine-tune Qwen3-4B-Base on four GB200 GPUs, with learning rate $10^{-6}$, 32 prompts per step, and 4,096-token prompt and 8,192-token response caps. A program gets reward one only if it passes all of its unit tests. We track validation accuracy with greedy decoding on 256 held-out AceCode tasks; it levels off after step 120 (Figure~\ref{fig:ace-teacher-current}, right), so we use step 120 as the final teacher and keep the rollouts up to that step. Over these steps, the teacher's avg@8 rises from 50.2\% to 74.9\% on HumanEval+ and from 48.8\% to 63.8\% on MBPP+.

\subsection{Datasets}
\label{app:data}
Table~\ref{tab:data-sizes} lists the size of every training set. To avoid test contamination in coding, we remove exact duplicates from AceCode, along with any task that shares a normalised 9-gram with a HumanEval+ or MBPP+ prompt.

\begin{table}[tbp]
\centering
\small
\setlength{\tabcolsep}{4.5pt}
\resizebox{\linewidth}{!}{%
\begin{tabular}{@{}llrrr@{}}
\toprule
Dataset & Selection & $\mathcal{T}_1$ & $\mathcal{T}_2$ & Coding \\
\midrule
\multicolumn{5}{@{}l}{\emph{Canonical distillation: a new generation pass from the final teacher}} \\
\Dsynth{} & correct responses from eight samples per training prompt & 57,380 & 64,767 & 16,702 \\
\Dsynth{}, one per prompt & one random correct response per prompt of \Dsynth{} & 8,739 & 8,980 & 16,702 \\
\midrule
\multicolumn{5}{@{}l}{\emph{Rollout by-products: no new teacher generation}} \\
\Dpool{} & every rollout saved during teacher RL & 239,592 & 122,880 & 61,440 \\
\DsftRand{} & one random correct rollout per solvable prompt & 10,897 & 6,460 & 3,252 \\
SFT, teacher-top1 or GRAPE & one correct rollout per solvable prompt, by likelihood & 10,897 & 6,460 & 3,252 \\
\DsftCurr{} & teacher-top1 correct rollout, easy $\to$ hard & 10,897 & 6,460 & 3,252 \\
\DsftAllp{} & teacher-top1 rollout on prompts with $\mathbb{E}[r](x)=1$ & 519 & 1,495 & 338 \\
\DdpoRand{} & one random correct--incorrect pair per mixed-success prompt & 10,378 & 4,965 & 2,914 \\
\DdpoCurr{} & teacher-top1 pair per mixed-success prompt, easy $\to$ hard & 10,378 & 4,965 & 2,914 \\
\bottomrule
\end{tabular}}
\caption{\textbf{Number of training examples in each dataset.} SFT rows are responses and DPO rows are correct--incorrect pairs. \Dsynth{} needs a new generation pass from the final teacher and keeps every correct response in maths and one per prompt in coding. The rollout datasets are all extracted from \Dpool{}, the completions saved during GRPO, so they need no new teacher generation; they differ in which prompts they use and how they pick responses (Table~\ref{tab:variants}).}
\label{tab:data-sizes}\label{tab:acecode-data-sizes}
\end{table}

\subsection{Student Training}
We train all students with TRL~\citep{vonwerra2020trl} and AdamW (Table~\ref{tab:train-hparams}). In maths, we train on eight H100 GPUs with a global batch of 256; DPO uses $\beta{=}0.1$ and DeepSpeed ZeRO-2. We tune learning rates only on the baselines: for \Dsynth{} on $\mathcal{T}_1$, we sweep $10^{-5}$ and $5\times10^{-5}$ with a constant or cosine schedule and keep the best run for each student. All rollout SFT runs then use $10^{-5}$ with a cosine schedule, and all DPO runs $5\times10^{-6}$ with a constant schedule. In coding, we train on four GB200 GPUs with a global batch of 64 and a maximum sequence length of 12,288 tokens.

\begin{table}[tbp]
\centering
\small
\setlength{\tabcolsep}{5pt}
\resizebox{\linewidth}{!}{%
\begin{tabular}{@{}lcccccc@{}}
\toprule
& \multicolumn{3}{c}{Maths SFT} & Maths DPO & \multicolumn{2}{c}{Coding} \\
\cmidrule(lr){2-4}\cmidrule(lr){6-7}
& \Dsynth{} & Rollouts & \DsftAllp{} & & SFT & DPO \\
\midrule
Learning rate & swept$^{*}$ & $10^{-5}$ cosine & $10^{-5}$ cosine & $5\times10^{-6}$ constant & $10^{-5}$ cosine & $5\times10^{-6}$ constant \\
Epochs & 1 & 2 & 6 & 2 & 2 & 2 \\
Max.\ sequence length & 4,096 & 4,096 & 4,096 & 2,048 & 12,288 & 12,288 \\
Global batch & 256 & 256 & 256 & 256 & 64 & 64 \\
DPO $\beta$ & -- & -- & -- & 0.1 & -- & 0.1 \\
\bottomrule
\end{tabular}}
\caption{\textbf{Student training hyperparameters for each method.} All runs use AdamW. $^{*}$For \Dsynth{} on $\mathcal{T}_1$, we sweep $10^{-5}$ and $5\times10^{-5}$ with a constant or cosine schedule and keep the best run for each student; \Dsynth{} on $\mathcal{T}_2$ uses $10^{-5}$ with a cosine schedule. For coding, LFM2.5-1.2B uses $5\times10^{-5}$ for \DsftRand{} and \Dsynth{}.}
\label{tab:train-hparams}
\end{table}

\subsection{Evaluation}
\label{app:inference}
We sample with vLLM~\citep{kwon2023efficient} at temperature $0.7$ (top-$p$ $0.95$ in coding) and report avg@8, the mean accuracy over eight samples per problem. We allow 3,072 generated tokens for MATH500, GSM8K and AMC 2023, 16,384 for AIME (7,168 for Gemma-2-2B, whose context is 8K tokens) and 8,192 in coding. We select checkpoints on MATH500 in maths and on the 256-task AceCode validation split in coding. For differences between methods, we report 95\% confidence intervals from a paired bootstrap over the test problems (100,000 resamples). For Gemma-2-2B, we use FlashAttention-2 in vLLM, since its FlashAttention-3 path lacks tanh soft-capping.

\FloatBarrier %
\section{Complete Results}
\label{app:full-results}
\begin{table}[H]
\centering
\small
\setlength{\tabcolsep}{1.2pt}
\begin{tabular}{@{}l c @{\hskip 4pt} ccccc @{\hskip 6pt} ccccc@{}}
\toprule
& & \multicolumn{5}{c}{$\mathcal{T}_1$: Qwen3-4B / Math12K} & \multicolumn{5}{c}{$\mathcal{T}_2$: Qwen3-8B / DeepScaler} \\
\cmidrule(lr){3-7}\cmidrule(l){8-12}
Student & Base & \Dsynth{} & \DsftRand{} & \DdpoRand{} & GRAPE & Ours & \Dsynth{} & \DsftRand{} & \DdpoRand{} & GRAPE & Ours \\
\midrule
Qwen2.5-1.5B & 35.3 & 51.0 & 46.3 & 52.2 & 49.8 & \textbf{54.6} & 46.1 & 40.8 & 42.0 & 44.8 & \textbf{47.4} \\
SmolLM3-3B & 55.1 & 57.8 & 54.6 & 58.8 & 59.1 & \textbf{59.9} & \textbf{64.6} & 62.3 & 57.2 & 61.0 & 63.1 \\
Gemma-2-2B & 3.8 & \textbf{23.2} & 18.6 & 7.2 & 21.8 & 17.1 & \textbf{20.6} & 13.6 & 5.0 & 17.7 & 16.6 \\
LFM2.5-1.2B & 37.9 & 49.7 & 47.5 & 42.8 & 50.9 & \textbf{51.0} & \textbf{57.4} & 51.1 & 41.5 & 52.5 & 54.1 \\
Qwen3-4B$^\dagger$ & 41.7 & 74.9 & 72.2 & \textbf{76.2} & -- & 75.5 & \textbf{78.4} & 77.8 & 75.0 & -- & \textbf{78.4} \\
\bottomrule
\end{tabular}

\vspace{6pt}
\begin{tabular}{@{}l ccccc @{\hskip 12pt} ccccc@{}}
\toprule
& \multicolumn{5}{c}{Coding: HumanEval+} & \multicolumn{5}{c}{Coding: MBPP+} \\
\cmidrule(lr){2-6}\cmidrule(l){7-11}
Student & Base & \Dsynth{} & \DsftRand{} & GRAPE & Ours & Base & \Dsynth{} & \DsftRand{} & GRAPE & Ours \\
\midrule
Qwen2.5-1.5B & 17.4 & \textbf{22.8} & 8.5 & 10.1 & 21.3 & 14.0 & 17.5 & 4.2 & 5.0 & \textbf{23.3} \\
SmolLM3-3B & 17.6 & \textbf{44.4} & 41.0 & 41.6 & 43.4 & 30.0 & \textbf{55.1} & 50.8 & 51.2 & 51.0 \\
LFM2.5-1.2B & 25.7 & 21.4 & \textbf{28.4} & 28.0 & \textbf{28.4} & 23.5 & 24.3 & 21.0 & \textbf{27.0} & 26.0 \\
Qwen3-4B$^\dagger$ & 50.2 & \textbf{76.8} & 73.9 & 74.8 & 76.4 & 48.8 & \textbf{65.5} & 64.4 & 62.8 & 65.0 \\
\bottomrule
\end{tabular}
\vspace{2pt}
\caption{\textbf{Main results on MATH500, HumanEval+ and MBPP+.} Avg@8 (\%) on MATH500 for both maths teachers (top) and on HumanEval+ and MBPP+ in coding (bottom). \Dsynth{} needs a new generation pass from the final teacher; every other column reuses the teacher's saved rollouts. Ours is \DtwoStage{}. GRAPE is SFT on the correct rollout that the base student finds most likely. Bold marks the best result in each row for each teacher or benchmark.
Tables~\ref{tab:comprehensive-t1}--\ref{tab:comprehensive-t2-continued} and~\ref{tab:acecode-multistudent} give the full results on every benchmark, with all rollout variants. $^\dagger$Same base model as the $\mathcal{T}_1$ and coding teachers.}
\label{tab:main-results}
\end{table}

\subsection{Mathematical Reasoning}
\label{app:full-math-results}\label{app:aime-rescore}
\begin{figure}[htbp]
    \centering
    \includegraphics[width=0.95\linewidth]{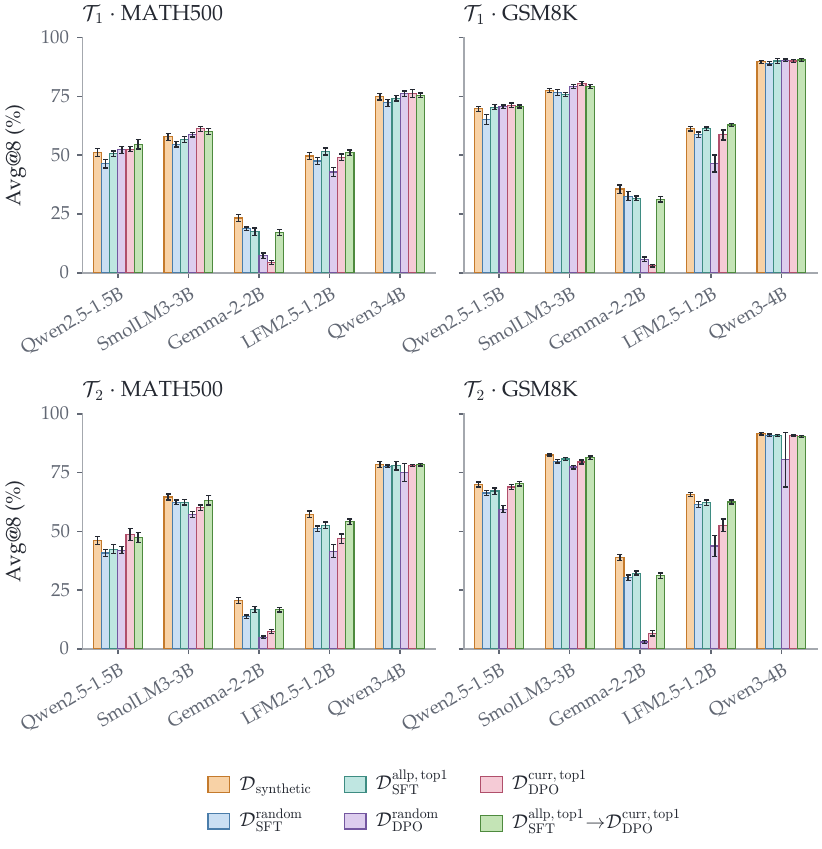}
    \caption{\textbf{Maths results for every method and student.} Avg@8 on MATH500 (left) and GSM8K (right) for $\mathcal{T}_1$ (top) and $\mathcal{T}_2$ (bottom); the MATH500 panels match Figure~\ref{fig:headline} in the main text. \Dsynth{} is generated on Math12K for both teachers.}
    \label{fig:math-gsm8k}
\end{figure}

Tables~\ref{tab:comprehensive-t1}--\ref{tab:comprehensive-t2-continued} report every maths method except GRAPE (Figures~\ref{fig:selector-crossbench-t1} and~\ref{fig:selector-crossbench-t2}) across all five benchmarks, and Figure~\ref{fig:math-gsm8k} shows MATH500 and GSM8K for every student. Averaged over both teachers, five students, and all five benchmarks weighted equally, our two-stage recipe (\DtwoStage{}) is 0.37 points below \Dsynth{} (95\% CI $[-0.82,+0.08]$, which includes zero), whereas naive rollout SFT (\DsftRand{}) is 1.95 points below ($[-2.33,-1.58]$).

\textbf{In maths, curated SFT data matches bigger SFT data.} Our SFT warm start (\DsftAllp{}) keeps the teacher-top1 response for only the consistently solved prompts ($\mathbb{E}[r](x)=1$). Despite using $21\times$ fewer prompts than \DsftRand{} on $\mathcal{T}_1$, it matches or outperforms \DsftRand{} for nearly every maths student and is on par with teacher-top1 SFT over all solvable prompts (\DsftCurr{}) on average (Tables~\ref{tab:comprehensive-t1} and~\ref{tab:comprehensive-t2}). We therefore use this split as the SFT warm start of our two-stage recipe, reserving the mixed-success prompts for DPO.

\begin{table}[H]
\centering
\small
\resizebox{\textwidth}{!}{%
\begin{tabular}{llccccc}
\toprule
Student & Data & AIME & AMC & MATH500 & GSM8K & Avg \\
\midrule
Qwen2.5-1.5B & Base & 0.004$\pm$0.008 & 0.134$\pm$0.030 & 0.353$\pm$0.012 & 0.546$\pm$0.032 & 0.259 \\
 & \Dsynth{} & 0.010$\pm$0.017 & \textbf{0.263$\pm$0.052} & 0.510$\pm$0.017 & 0.697$\pm$0.011 & 0.370 \\
 & \DsftRand{} & 0.019$\pm$0.011 & 0.200$\pm$0.057 & 0.463$\pm$0.019 & 0.652$\pm$0.021 & 0.334 \\
 & \DdpoRand{} & 0.017$\pm$0.012 & 0.256$\pm$0.029 & 0.522$\pm$0.015 & 0.707$\pm$0.007 & 0.375 \\
 & \DsftAllp{} & 0.015$\pm$0.017 & 0.247$\pm$0.034 & 0.506$\pm$0.011 & 0.704$\pm$0.011 & 0.368 \\
 & \DsftCurr{} & \textbf{0.023$\pm$0.015} & \textbf{0.263$\pm$0.050} & 0.513$\pm$0.013 & 0.691$\pm$0.006 & 0.372 \\
 & \DdpoCurr{} & 0.013$\pm$0.017 & 0.241$\pm$0.038 & 0.525$\pm$0.011 & 0.712$\pm$0.009 & 0.373 \\
 & \DtwoStage{} & 0.013$\pm$0.019 & 0.234$\pm$0.053 & \textbf{0.546$\pm$0.020} & 0.707$\pm$0.007 & 0.375 \\
 & \DsftRand{}$\to$\DdpoRand{} & 0.021$\pm$0.021 & 0.259$\pm$0.063 & 0.521$\pm$0.012 & \textbf{0.719$\pm$0.009} & \textbf{0.380} \\
\cmidrule(lr){2-7}
 & SFT, teacher-top1 (shuffled) & 0.015$\pm$0.015 & 0.194$\pm$0.032 & 0.499$\pm$0.018 & 0.683$\pm$0.010 & 0.348 \\
 & DPO, teacher-top1 (shuffled) & 0.017$\pm$0.013 & 0.238$\pm$0.050 & 0.544$\pm$0.014 & \textbf{0.719$\pm$0.008} & 0.379 \\
\midrule
SmolLM3-3B-Base & Base & 0.019$\pm$0.018 & 0.281$\pm$0.055 & 0.551$\pm$0.025 & 0.763$\pm$0.011 & 0.404 \\
 & \Dsynth{} & 0.023$\pm$0.021 & 0.294$\pm$0.065 & 0.578$\pm$0.015 & 0.774$\pm$0.009 & 0.417 \\
 & \DsftRand{} & 0.025$\pm$0.020 & 0.322$\pm$0.067 & 0.546$\pm$0.012 & 0.767$\pm$0.012 & 0.415 \\
 & \DdpoRand{} & 0.031$\pm$0.026 & 0.322$\pm$0.021 & 0.588$\pm$0.010 & 0.793$\pm$0.008 & 0.433 \\
 & \DsftAllp{} & 0.021$\pm$0.015 & 0.300$\pm$0.048 & 0.567$\pm$0.012 & 0.759$\pm$0.008 & 0.412 \\
 & \DsftCurr{} & 0.027$\pm$0.020 & 0.284$\pm$0.042 & 0.565$\pm$0.012 & 0.764$\pm$0.009 & 0.410 \\
 & \DdpoCurr{} & 0.033$\pm$0.023 & \textbf{0.350$\pm$0.042} & \textbf{0.612$\pm$0.011} & \textbf{0.804$\pm$0.009} & \textbf{0.450} \\
 & \DtwoStage{} & \textbf{0.048$\pm$0.018} & 0.341$\pm$0.073 & 0.599$\pm$0.013 & 0.792$\pm$0.008 & 0.445 \\
 & \DsftRand{}$\to$\DdpoRand{} & 0.046$\pm$0.024 & \textbf{0.350$\pm$0.079} & 0.597$\pm$0.012 & 0.798$\pm$0.009 & 0.448 \\
\cmidrule(lr){2-7}
 & SFT, teacher-top1 (shuffled) & 0.029$\pm$0.020 & 0.303$\pm$0.028 & 0.570$\pm$0.012 & 0.766$\pm$0.009 & 0.417 \\
 & DPO, teacher-top1 (shuffled) & 0.038$\pm$0.020 & 0.347$\pm$0.063 & 0.605$\pm$0.019 & 0.799$\pm$0.011 & 0.447 \\
\midrule
Gemma-2-2B & Base & 0.000$\pm$0.000 & 0.013$\pm$0.019 & 0.038$\pm$0.011 & 0.020$\pm$0.006 & 0.017 \\
 & \Dsynth{} & 0.000$\pm$0.000 & \textbf{0.094$\pm$0.032} & \textbf{0.232$\pm$0.014} & \textbf{0.356$\pm$0.018} & \textbf{0.171} \\
 & \DsftRand{} & 0.004$\pm$0.008 & 0.044$\pm$0.029 & 0.186$\pm$0.007 & 0.326$\pm$0.019 & 0.140 \\
 & \DdpoRand{} & 0.000$\pm$0.000 & 0.028$\pm$0.034 & 0.072$\pm$0.012 & 0.057$\pm$0.010 & 0.039 \\
 & \DsftAllp{} & 0.004$\pm$0.008 & 0.056$\pm$0.037 & 0.174$\pm$0.015 & 0.317$\pm$0.010 & 0.138 \\
 & \DsftCurr{} & 0.002$\pm$0.006 & 0.069$\pm$0.026 & 0.208$\pm$0.014 & 0.348$\pm$0.005 & 0.157 \\
 & \DdpoCurr{} & 0.000$\pm$0.000 & 0.003$\pm$0.009 & 0.043$\pm$0.009 & 0.030$\pm$0.006 & 0.019 \\
 & \DtwoStage{} & 0.000$\pm$0.000 & 0.059$\pm$0.033 & 0.171$\pm$0.012 & 0.313$\pm$0.011 & 0.136 \\
 & \DsftRand{}$\to$\DdpoRand{} & 0.000$\pm$0.000 & 0.084$\pm$0.048 & 0.189$\pm$0.015 & 0.324$\pm$0.014 & 0.149 \\
\cmidrule(lr){2-7}
 & SFT, teacher-top1 (shuffled) & \textbf{0.006$\pm$0.010} & 0.062$\pm$0.035 & 0.202$\pm$0.009 & 0.339$\pm$0.012 & 0.152 \\
 & DPO, teacher-top1 (shuffled) & 0.000$\pm$0.000 & 0.022$\pm$0.016 & 0.043$\pm$0.009 & 0.031$\pm$0.004 & 0.024 \\
\bottomrule
\end{tabular}}
\caption{\textbf{Maths accuracy of every method and student on all benchmarks ($\mathcal{T}_1$).} Avg@8 $\pm$ standard deviation over the eight samples, at temperature $0.7$. AIME averages the 2024 and 2025 scores; Avg averages MATH500, GSM8K, AMC, and the combined AIME score. Bold marks the highest value per student and column. The rows below the rule are the shuffled teacher-top1 runs of Figure~\ref{fig:selection}. \DsftRand{}$\to$\DdpoRand{} runs DPO on random pairs from the random-SFT checkpoint.}
\label{tab:comprehensive-t1}
\end{table}

\clearpage
\begin{table}[H]
\centering
\small
\resizebox{\textwidth}{!}{%
\begin{tabular}{llccccc}
\toprule
Student & Data & AIME & AMC & MATH500 & GSM8K & Avg \\
\midrule
LFM2.5-1.2B-Base & Base & 0.021$\pm$0.020 & 0.156$\pm$0.044 & 0.379$\pm$0.037 & 0.385$\pm$0.037 & 0.235 \\
 & \Dsynth{} & 0.010$\pm$0.011 & 0.234$\pm$0.048 & 0.497$\pm$0.016 & 0.613$\pm$0.009 & 0.339 \\
 & \DsftRand{} & 0.013$\pm$0.012 & 0.238$\pm$0.057 & 0.475$\pm$0.015 & 0.587$\pm$0.012 & 0.328 \\
 & \DdpoRand{} & 0.019$\pm$0.025 & 0.194$\pm$0.037 & 0.428$\pm$0.018 & 0.464$\pm$0.037 & 0.276 \\
 & \DsftAllp{} & \textbf{0.027$\pm$0.018} & 0.244$\pm$0.046 & \textbf{0.516$\pm$0.015} & 0.613$\pm$0.007 & 0.350 \\
 & \DsftCurr{} & 0.023$\pm$0.021 & 0.253$\pm$0.043 & 0.479$\pm$0.014 & 0.589$\pm$0.015 & 0.336 \\
 & \DdpoCurr{} & 0.021$\pm$0.019 & 0.241$\pm$0.048 & 0.491$\pm$0.015 & 0.586$\pm$0.021 & 0.335 \\
 & \DtwoStage{} & 0.025$\pm$0.019 & 0.213$\pm$0.071 & 0.510$\pm$0.011 & \textbf{0.629$\pm$0.006} & 0.344 \\
 & \DsftRand{}$\to$\DdpoRand{} & 0.021$\pm$0.015 & 0.206$\pm$0.040 & 0.486$\pm$0.014 & 0.581$\pm$0.010 & 0.324 \\
\cmidrule(lr){2-7}
 & SFT, teacher-top1 (shuffled) & 0.023$\pm$0.017 & \textbf{0.266$\pm$0.060} & 0.504$\pm$0.017 & 0.628$\pm$0.009 & \textbf{0.355} \\
 & DPO, teacher-top1 (shuffled) & 0.019$\pm$0.018 & 0.191$\pm$0.027 & 0.459$\pm$0.018 & 0.543$\pm$0.022 & 0.303 \\
\midrule
Qwen3-4B-Base & Base & 0.069$\pm$0.032 & 0.216$\pm$0.072 & 0.417$\pm$0.062 & 0.488$\pm$0.050 & 0.298 \\
 & \Dsynth{} & 0.079$\pm$0.028 & 0.534$\pm$0.044 & 0.749$\pm$0.014 & 0.897$\pm$0.006 & 0.565 \\
 & \DsftRand{} & 0.100$\pm$0.029 & 0.509$\pm$0.055 & 0.722$\pm$0.016 & 0.890$\pm$0.007 & 0.556 \\
 & \DdpoRand{} & 0.110$\pm$0.028 & 0.519$\pm$0.046 & 0.762$\pm$0.011 & 0.904$\pm$0.004 & 0.574 \\
 & \DsftAllp{} & 0.083$\pm$0.028 & 0.516$\pm$0.080 & 0.743$\pm$0.012 & 0.901$\pm$0.010 & 0.561 \\
 & \DsftCurr{} & 0.098$\pm$0.033 & 0.534$\pm$0.046 & 0.747$\pm$0.008 & 0.901$\pm$0.004 & 0.570 \\
 & \DdpoCurr{} & 0.090$\pm$0.034 & 0.509$\pm$0.035 & 0.762$\pm$0.017 & 0.901$\pm$0.006 & 0.566 \\
 & \DtwoStage{} & 0.108$\pm$0.026 & 0.544$\pm$0.046 & 0.755$\pm$0.010 & 0.905$\pm$0.005 & 0.578 \\
 & \DsftRand{}$\to$\DdpoRand{} & 0.090$\pm$0.023 & 0.522$\pm$0.043 & 0.740$\pm$0.012 & 0.902$\pm$0.003 & 0.563 \\
\cmidrule(lr){2-7}
 & DPO, teacher-top1 (shuffled) & \textbf{0.131$\pm$0.026} & \textbf{0.550$\pm$0.033} & \textbf{0.786$\pm$0.008} & \textbf{0.906$\pm$0.005} & \textbf{0.593} \\
\bottomrule
\end{tabular}}
\caption{\textbf{Maths accuracy of every method and student on all benchmarks ($\mathcal{T}_1$, continued from Table~\ref{tab:comprehensive-t1}):} LFM2.5-1.2B-Base and Qwen3-4B-Base.}
\label{tab:comprehensive-t1-continued}
\end{table}

\begin{table}[H]
\centering
\small
\resizebox{\textwidth}{!}{%
\begin{tabular}{llccccc}
\toprule
Student & Data & AIME & AMC & MATH500 & GSM8K & Avg \\
\midrule
Qwen2.5-1.5B & Base & 0.004$\pm$0.008 & 0.134$\pm$0.030 & 0.353$\pm$0.012 & 0.546$\pm$0.032 & 0.259 \\
 & \Dsynth{} & 0.019$\pm$0.015 & 0.231$\pm$0.066 & 0.461$\pm$0.016 & 0.699$\pm$0.011 & 0.353 \\
 & \DsftRand{} & 0.013$\pm$0.014 & 0.172$\pm$0.074 & 0.408$\pm$0.014 & 0.663$\pm$0.010 & 0.314 \\
 & \DdpoRand{} & 0.006$\pm$0.009 & 0.172$\pm$0.057 & 0.420$\pm$0.014 & 0.595$\pm$0.015 & 0.298 \\
 & \DsftAllp{} & 0.010$\pm$0.011 & 0.172$\pm$0.041 & 0.424$\pm$0.019 & 0.672$\pm$0.014 & 0.320 \\
 & \DsftCurr{} & 0.008$\pm$0.009 & 0.191$\pm$0.038 & 0.428$\pm$0.010 & 0.675$\pm$0.007 & 0.325 \\
 & \DdpoCurr{} & 0.021$\pm$0.016 & 0.194$\pm$0.050 & \textbf{0.486$\pm$0.025} & 0.689$\pm$0.009 & 0.347 \\
 & \DtwoStage{} & 0.019$\pm$0.017 & 0.219$\pm$0.032 & 0.474$\pm$0.021 & \textbf{0.703$\pm$0.009} & 0.354 \\
 & \DsftRand{}$\to$\DdpoRand{} & \textbf{0.029$\pm$0.021} & \textbf{0.247$\pm$0.039} & 0.469$\pm$0.016 & 0.690$\pm$0.011 & \textbf{0.359} \\
\cmidrule(lr){2-7}
 & SFT, teacher-top1 (shuffled) & 0.010$\pm$0.011 & 0.194$\pm$0.040 & 0.407$\pm$0.014 & 0.673$\pm$0.006 & 0.321 \\
 & DPO, teacher-top1 (shuffled) & 0.023$\pm$0.018 & 0.222$\pm$0.049 & 0.483$\pm$0.019 & 0.672$\pm$0.012 & 0.350 \\
\midrule
SmolLM3-3B-Base & Base & 0.019$\pm$0.018 & 0.281$\pm$0.055 & 0.551$\pm$0.025 & 0.763$\pm$0.011 & 0.404 \\
 & \Dsynth{} & 0.075$\pm$0.028 & \textbf{0.387$\pm$0.030} & \textbf{0.646$\pm$0.013} & \textbf{0.825$\pm$0.004} & \textbf{0.483} \\
 & \DsftRand{} & 0.079$\pm$0.033 & 0.369$\pm$0.042 & 0.623$\pm$0.010 & 0.799$\pm$0.008 & 0.467 \\
 & \DdpoRand{} & 0.029$\pm$0.017 & 0.353$\pm$0.049 & 0.572$\pm$0.013 & 0.774$\pm$0.008 & 0.432 \\
 & \DsftAllp{} & 0.088$\pm$0.026 & 0.356$\pm$0.026 & 0.623$\pm$0.011 & 0.808$\pm$0.008 & 0.469 \\
 & \DsftCurr{} & 0.073$\pm$0.033 & 0.359$\pm$0.033 & 0.619$\pm$0.009 & 0.798$\pm$0.007 & 0.462 \\
 & \DdpoCurr{} & 0.029$\pm$0.019 & 0.344$\pm$0.070 & 0.601$\pm$0.012 & 0.795$\pm$0.008 & 0.442 \\
 & \DtwoStage{} & \textbf{0.092$\pm$0.025} & 0.362$\pm$0.044 & 0.631$\pm$0.020 & 0.814$\pm$0.007 & 0.475 \\
 & \DsftRand{}$\to$\DdpoRand{} & 0.081$\pm$0.038 & 0.378$\pm$0.047 & 0.632$\pm$0.017 & 0.812$\pm$0.011 & 0.476 \\
\cmidrule(lr){2-7}
 & SFT, teacher-top1 (shuffled) & 0.083$\pm$0.022 & 0.359$\pm$0.035 & 0.621$\pm$0.015 & 0.800$\pm$0.009 & 0.466 \\
 & DPO, teacher-top1 (shuffled) & 0.038$\pm$0.019 & 0.316$\pm$0.057 & 0.583$\pm$0.014 & 0.792$\pm$0.005 & 0.432 \\
\midrule
Gemma-2-2B & Base & 0.000$\pm$0.000 & 0.013$\pm$0.019 & 0.038$\pm$0.011 & 0.020$\pm$0.006 & 0.017 \\
 & \Dsynth{} & \textbf{0.004$\pm$0.008} & \textbf{0.069$\pm$0.026} & \textbf{0.206$\pm$0.013} & \textbf{0.388$\pm$0.013} & \textbf{0.167} \\
 & \DsftRand{} & 0.002$\pm$0.006 & 0.041$\pm$0.033 & 0.136$\pm$0.007 & 0.303$\pm$0.012 & 0.120 \\
 & \DdpoRand{} & 0.000$\pm$0.000 & 0.013$\pm$0.019 & 0.050$\pm$0.005 & 0.030$\pm$0.007 & 0.023 \\
 & \DsftAllp{} & \textbf{0.004$\pm$0.008} & 0.031$\pm$0.035 & 0.167$\pm$0.013 & 0.322$\pm$0.009 & 0.131 \\
 & \DsftCurr{} & 0.002$\pm$0.006 & 0.062$\pm$0.023 & 0.169$\pm$0.017 & 0.330$\pm$0.015 & 0.141 \\
 & \DdpoCurr{} & 0.000$\pm$0.000 & 0.028$\pm$0.021 & 0.074$\pm$0.007 & 0.067$\pm$0.012 & 0.042 \\
 & \DtwoStage{} & 0.000$\pm$0.000 & 0.056$\pm$0.032 & 0.166$\pm$0.010 & 0.312$\pm$0.012 & 0.134 \\
 & \DsftRand{}$\to$\DdpoRand{} & 0.002$\pm$0.006 & 0.041$\pm$0.030 & 0.144$\pm$0.014 & 0.304$\pm$0.011 & 0.123 \\
\cmidrule(lr){2-7}
 & SFT, teacher-top1 (shuffled) & 0.000$\pm$0.000 & 0.038$\pm$0.030 & 0.161$\pm$0.015 & 0.318$\pm$0.015 & 0.129 \\
 & DPO, teacher-top1 (shuffled) & 0.000$\pm$0.000 & 0.009$\pm$0.019 & 0.044$\pm$0.008 & 0.028$\pm$0.008 & 0.020 \\
\bottomrule
\end{tabular}}
\caption{\textbf{Maths accuracy of every method and student on all benchmarks ($\mathcal{T}_2$).} Avg@8 $\pm$ standard deviation over the eight samples, at temperature $0.7$. AIME averages the 2024 and 2025 scores; Avg averages MATH500, GSM8K, AMC, and the combined AIME score. Bold marks the highest value per student and column. The rows below the rule are the shuffled teacher-top1 runs of Figure~\ref{fig:selection}. Here \Dsynth{} regenerates on Math12K, the training split of MATH500, while every rollout dataset reuses the DeepScaler rollouts of $\mathcal{T}_2$.}
\label{tab:comprehensive-t2}
\end{table}

\clearpage
\begin{table}[H]
\centering
\small
\resizebox{\textwidth}{!}{%
\begin{tabular}{llccccc}
\toprule
Student & Data & AIME & AMC & MATH500 & GSM8K & Avg \\
\midrule
LFM2.5-1.2B-Base & Base & 0.021$\pm$0.020 & 0.156$\pm$0.044 & 0.379$\pm$0.037 & 0.385$\pm$0.037 & 0.235 \\
 & \Dsynth{} & \textbf{0.058$\pm$0.019} & \textbf{0.284$\pm$0.078} & \textbf{0.574$\pm$0.013} & \textbf{0.658$\pm$0.008} & \textbf{0.393} \\
 & \DsftRand{} & 0.048$\pm$0.023 & 0.266$\pm$0.038 & 0.511$\pm$0.011 & 0.614$\pm$0.012 & 0.360 \\
 & \DdpoRand{} & 0.017$\pm$0.015 & 0.184$\pm$0.030 & 0.415$\pm$0.028 & 0.438$\pm$0.044 & 0.264 \\
 & \DsftAllp{} & 0.054$\pm$0.028 & 0.247$\pm$0.034 & 0.526$\pm$0.014 & 0.621$\pm$0.012 & 0.362 \\
 & \DsftCurr{} & 0.052$\pm$0.024 & 0.253$\pm$0.056 & 0.522$\pm$0.011 & 0.613$\pm$0.012 & 0.360 \\
 & \DdpoCurr{} & 0.027$\pm$0.023 & 0.216$\pm$0.053 & 0.469$\pm$0.020 & 0.527$\pm$0.026 & 0.310 \\
 & \DtwoStage{} & 0.042$\pm$0.011 & 0.244$\pm$0.029 & 0.541$\pm$0.012 & 0.625$\pm$0.009 & 0.363 \\
 & \DsftRand{}$\to$\DdpoRand{} & 0.056$\pm$0.029 & 0.250$\pm$0.065 & 0.513$\pm$0.019 & 0.611$\pm$0.012 & 0.358 \\
\cmidrule(lr){2-7}
 & SFT, teacher-top1 (shuffled) & 0.050$\pm$0.032 & 0.256$\pm$0.044 & 0.522$\pm$0.018 & 0.619$\pm$0.010 & 0.362 \\
 & DPO, teacher-top1 (shuffled) & 0.019$\pm$0.018 & 0.166$\pm$0.042 & 0.422$\pm$0.032 & 0.475$\pm$0.036 & 0.270 \\
\midrule
Qwen3-4B-Base & Base & 0.069$\pm$0.032 & 0.216$\pm$0.072 & 0.417$\pm$0.062 & 0.488$\pm$0.050 & 0.298 \\
 & \Dsynth{} & 0.213$\pm$0.034 & 0.534$\pm$0.048 & \textbf{0.784$\pm$0.013} & \textbf{0.915$\pm$0.005} & \textbf{0.611} \\
 & \DsftRand{} & 0.185$\pm$0.034 & 0.497$\pm$0.062 & 0.778$\pm$0.005 & 0.909$\pm$0.005 & 0.592 \\
 & \DdpoRand{} & 0.169$\pm$0.030 & 0.459$\pm$0.019 & 0.750$\pm$0.039 & 0.804$\pm$0.116 & 0.546 \\
 & \DsftAllp{} & 0.204$\pm$0.045 & 0.528$\pm$0.025 & 0.779$\pm$0.017 & 0.907$\pm$0.005 & 0.605 \\
 & \DsftCurr{} & 0.188$\pm$0.042 & 0.522$\pm$0.045 & 0.781$\pm$0.008 & 0.910$\pm$0.005 & 0.600 \\
 & \DdpoCurr{} & 0.185$\pm$0.028 & 0.519$\pm$0.053 & 0.780$\pm$0.005 & 0.908$\pm$0.002 & 0.598 \\
 & \DtwoStage{} & 0.200$\pm$0.030 & \textbf{0.537$\pm$0.052} & \textbf{0.784$\pm$0.006} & 0.904$\pm$0.004 & 0.606 \\
 & \DsftRand{}$\to$\DdpoRand{} & \textbf{0.217$\pm$0.029} & 0.494$\pm$0.050 & 0.780$\pm$0.013 & 0.904$\pm$0.007 & 0.599 \\
\cmidrule(lr){2-7}
 & SFT, teacher-top1 (shuffled) & 0.190$\pm$0.024 & 0.500$\pm$0.061 & 0.778$\pm$0.007 & 0.906$\pm$0.009 & 0.593 \\
 & DPO, teacher-top1 (shuffled) & 0.177$\pm$0.031 & 0.500$\pm$0.046 & 0.782$\pm$0.023 & 0.871$\pm$0.068 & 0.583 \\
\bottomrule
\end{tabular}}
\caption{\textbf{Maths accuracy of every method and student on all benchmarks ($\mathcal{T}_2$, continued from Table~\ref{tab:comprehensive-t2}):} LFM2.5-1.2B-Base and Qwen3-4B-Base.}
\label{tab:comprehensive-t2-continued}
\end{table}

\subsection{Longer DPO Training Sequences}
\label{app:dpo-length}
\begin{figure}[htbp]
    \centering
    \includegraphics[width=\linewidth]{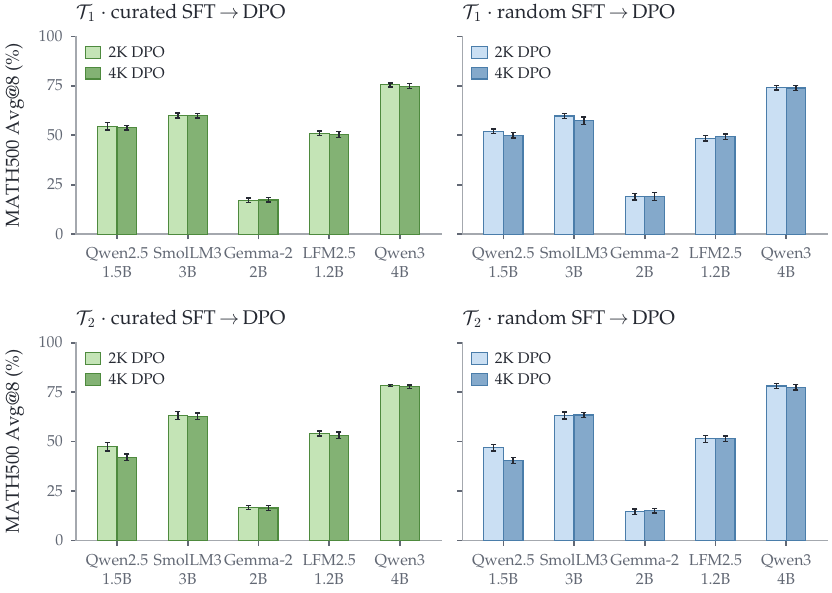}
    \caption{\textbf{Two-stage training with 2,048- and 4,096-token DPO.} MATH500 avg@8 for curated (\DtwoStage{}, left) and random (\DsftRand{}$\to$\DdpoRand{}, right) two-stage training under $\mathcal{T}_1$ (top) and $\mathcal{T}_2$ (bottom), with the same SFT warm starts. Error bars show one standard deviation over the eight samples.}
    \label{fig:dpo-length-twostage}
\end{figure}

\begin{figure}[htbp]
    \centering
    \includegraphics[width=\linewidth]{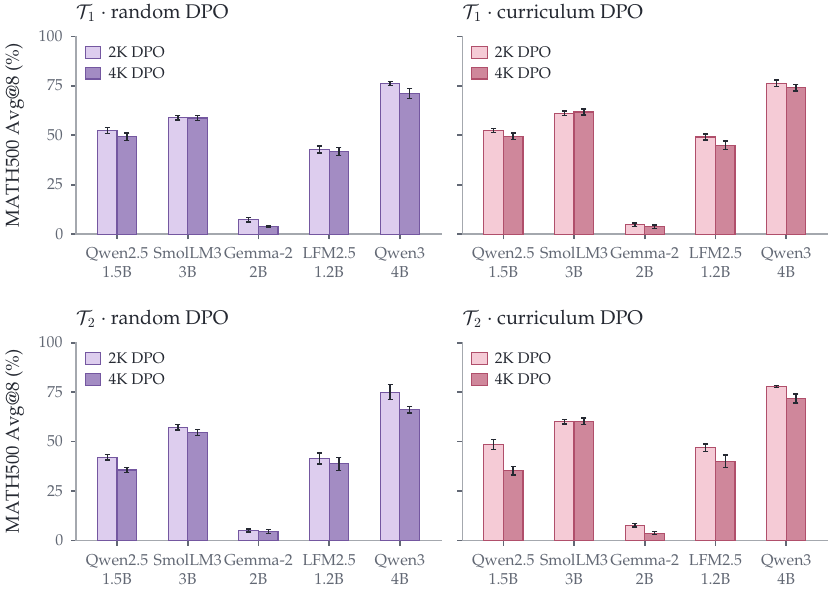}
    \caption{\textbf{Standalone DPO with 2,048- and 4,096-token training.} MATH500 avg@8 for \DdpoRand{} (left) and \DdpoCurr{} (right) under $\mathcal{T}_1$ (top) and $\mathcal{T}_2$ (bottom). Error bars show one standard deviation over the eight samples.}
    \label{fig:dpo-length-standalone}
\end{figure}

We also rerun every maths DPO experiment (standalone random and curriculum DPO, and both two-stage variants) with a 4,096-token maximum sequence length instead of 2,048 (Figures~\ref{fig:dpo-length-twostage} and~\ref{fig:dpo-length-standalone}). We do not run a hyperparameter sweep for the 4,096-token runs; we use the same hyperparameters as the 2,048-token runs and the same SFT checkpoints as warm starts. Standalone DPO regresses at 4,096 tokens, most for the Qwen students and LFM2.5-1.2B, while the two-stage results stay similar except for Qwen2.5-1.5B under $\mathcal{T}_2$.

\FloatBarrier

\subsection{One Synthetic Response per Prompt}
\label{app:synthetic-size-control}
In maths, \Dsynth{} keeps every correct response from the eight samples per prompt, whereas our rollout datasets keep one response per prompt. To isolate the effect of dataset size, we subsample one random correct response for each of the 8,739 solved prompts in the $\mathcal{T}_1$ \Dsynth{} pool and train under identical settings (Table~\ref{tab:synthetic-size-control}). For four of the five students, using one response per prompt lies within $1.3$ points of full \Dsynth{} on MATH500; Gemma-2-2B is the exception, dropping from $23.2\%$ to $18.0\%$.

\begin{table}[H]
\centering
\small
\setlength{\tabcolsep}{3.8pt}
\begin{tabular}{@{}lrrrr@{}}
\toprule
& \DsftRand{} & \Dsynth{}, one per prompt & \Dsynth{}, all correct & \DtwoStage{} \\
Rows & 10,897 & 8,739 & 57,380 & 519 + 10,378 \\
\midrule
Qwen2.5-1.5B & 46.3 & 51.6 & 51.0 & 54.6 \\
SmolLM3-3B & 54.6 & 57.3 & 57.8 & 59.9 \\
Gemma-2-2B & 18.6 & 18.0 & 23.2 & 17.1 \\
LFM2.5-1.2B & 47.5 & 48.4 & 49.7 & 51.0 \\
Qwen3-4B & 72.2 & 75.7 & 74.9 & 75.5 \\
\bottomrule
\end{tabular}
\caption{\textbf{Canonical distillation with one versus all correct synthetic responses per prompt.} MATH500 avg@8 (\%) on $\mathcal{T}_1$. The one-per-prompt \Dsynth{} keeps one random correct response for each prompt of \Dsynth{} and uses the same training settings.}
\label{tab:synthetic-size-control}
\end{table}

\subsection{Python Coding}
\label{app:coding-details}\label{app:nonmath-audits}
Table~\ref{tab:acecode-multistudent} reports the coding methods for the four students, and Figure~\ref{fig:headline-code} in the main text shows the main comparison. Across the eight student--benchmark cells, the two-stage recipe improves on the base model in every cell and is on par with \Dsynth{} on average ($+0.89$ points, 95\% CI $[-0.12,+1.91]$), whereas naive SFT falls 4.44 points below \Dsynth{} ($[-5.37,-3.53]$).

\begin{table}[!t]
\centering
\small
\setlength{\tabcolsep}{2.8pt}
\resizebox{\textwidth}{!}{%
\begin{tabular}{llrrrrrrr}
\toprule
Benchmark & Student & Base & \Dsynth{} & \DsftRand{} & \DsftAllp{} &
SFT, teacher-top1 & SFT, GRAPE & \DtwoStage{} \\
\midrule
HumanEval+ & Qwen2.5-1.5B & $17.38{\pm}1.46$ & $\mathbf{22.79{\pm}1.87}$ & $8.54{\pm}1.53$ & $18.06{\pm}2.71$ & $10.75{\pm}2.42$ & $10.14{\pm}2.01$ & $21.34{\pm}1.56$ \\
 & SmolLM3-3B & $17.61{\pm}3.10$ & $\mathbf{44.36{\pm}2.37}$ & $41.01{\pm}2.82$ & $29.34{\pm}2.51$ & $39.02{\pm}2.71$ & $41.62{\pm}2.18$ & $43.45{\pm}1.69$ \\
 & LFM2.5-1.2B & $25.69{\pm}2.17$ & $21.42{\pm}2.58$ & $28.43{\pm}2.75$ & $24.24{\pm}1.41$ & $\mathbf{30.56{\pm}1.61}$ & $28.05{\pm}2.06$ & $28.43{\pm}2.59$ \\
 & Qwen3-4B-Base & $50.15{\pm}2.54$ & $\mathbf{76.83{\pm}1.81}$ & $73.86{\pm}2.02$ & $75.84{\pm}2.39$ & $76.30{\pm}3.25$ & $74.77{\pm}2.95$ & $76.37{\pm}1.16$ \\
\midrule
MBPP+ & Qwen2.5-1.5B & $13.99{\pm}1.07$ & $17.49{\pm}2.11$ & $4.17{\pm}0.88$ & $18.06{\pm}1.68$ & $5.46{\pm}1.09$ & $4.99{\pm}0.81$ & $\mathbf{23.31{\pm}1.45}$ \\
 & SmolLM3-3B & $29.96{\pm}2.26$ & $\mathbf{55.06{\pm}1.22}$ & $50.83{\pm}1.77$ & $47.16{\pm}1.86$ & $49.83{\pm}1.41$ & $51.19{\pm}2.03$ & $50.96{\pm}1.80$ \\
 & LFM2.5-1.2B & $23.48{\pm}2.53$ & $24.27{\pm}0.85$ & $21.00{\pm}2.15$ & $23.21{\pm}2.41$ & $\mathbf{27.94{\pm}1.08}$ & $27.02{\pm}1.34$ & $25.96{\pm}1.52$ \\
 & Qwen3-4B-Base & $48.78{\pm}2.04$ & $\mathbf{65.54{\pm}1.10}$ & $64.38{\pm}1.88$ & $64.42{\pm}1.62$ & $64.15{\pm}0.51$ & $62.80{\pm}0.71$ & $65.05{\pm}0.92$ \\
\bottomrule
\end{tabular}}
\caption{\textbf{Coding accuracy of the reported methods on HumanEval+ and MBPP+.} Avg@8 (\%) $\pm$ standard deviation over the eight samples, at temperature $0.7$. Bold marks the best trained model in each row.}
\label{tab:acecode-multistudent}
\end{table}

\FloatBarrier %
\section{Response Selection}
This section gives the data behind Section~\ref{sec:q2}: the selection and ordering controls, the comparison with GRAPE, and the RL steps that the selected responses come from.
\label{app:selection}

\subsection{Selection and Ordering Controls}
Table~\ref{tab:selection-order} gives the MATH500 values behind Figure~\ref{fig:selection} and the ordering comparison of Section~\ref{sec:decoupling}.

\begin{table}[H]
\centering
\small
\setlength{\tabcolsep}{4.5pt}
\begin{tabular}{lrrrrrr}
\toprule
& \multicolumn{3}{c}{SFT, all solvable prompts} & \multicolumn{3}{c}{DPO, mixed-success prompts} \\
\cmidrule(lr){2-4}\cmidrule(lr){5-7}
Student & random & top1 & top1 & random & top1 & top1 \\
 & shuffled & shuffled & curriculum & shuffled & shuffled & curriculum \\
\midrule
\multicolumn{7}{l}{\emph{$\mathcal{T}_1$}} \\
Qwen2.5-1.5B & 46.3 & 49.9 & 51.3 & 52.2 & 54.4 & 52.5 \\
SmolLM3-3B & 54.6 & 57.0 & 56.5 & 58.8 & 60.5 & 61.2 \\
Gemma-2-2B & 18.6 & 20.2 & 20.8 & 7.2 & 4.3 & 4.3 \\
LFM2.5-1.2B & 47.5 & 50.4 & 47.9 & 42.8 & 45.9 & 49.1 \\
Qwen3-4B & 72.2 & -- & 74.7 & 76.2 & 78.6 & 76.2 \\
\midrule
\multicolumn{7}{l}{\emph{$\mathcal{T}_2$}} \\
Qwen2.5-1.5B & 40.8 & 40.7 & 42.8 & 42.0 & 48.3 & 48.6 \\
SmolLM3-3B & 62.3 & 62.1 & 61.9 & 57.2 & 58.3 & 60.1 \\
Gemma-2-2B & 13.6 & 16.1 & 16.9 & 5.0 & 4.4 & 7.4 \\
LFM2.5-1.2B & 51.1 & 52.2 & 52.2 & 41.5 & 42.2 & 46.9 \\
Qwen3-4B & 77.8 & 77.8 & 78.1 & 75.0 & 78.2 & 78.0 \\
\midrule
\emph{Mean change} & & +1.5 (6/9) & +0.2 (5/9) & & +1.7 (8/10) & +0.9 (6/10) \\
\bottomrule
\end{tabular}
\caption{\textbf{Isolating the effects of response selection and prompt ordering.} MATH500 avg@8 (\%). Within each stage, all three runs use the same prompts. We compare random and teacher-top1 selection with the prompts shuffled, and shuffled and easy $\to$ hard order with teacher-top1 selection. The last row averages each change over all settings, with the number of settings that improve in parentheses.}
\label{tab:selection-order}
\end{table}

\subsection{Random, Teacher-Top1 and GRAPE}
\label{app:selector-cohorts}
\begin{figure}[H]
\centering
\includegraphics[width=0.95\linewidth]{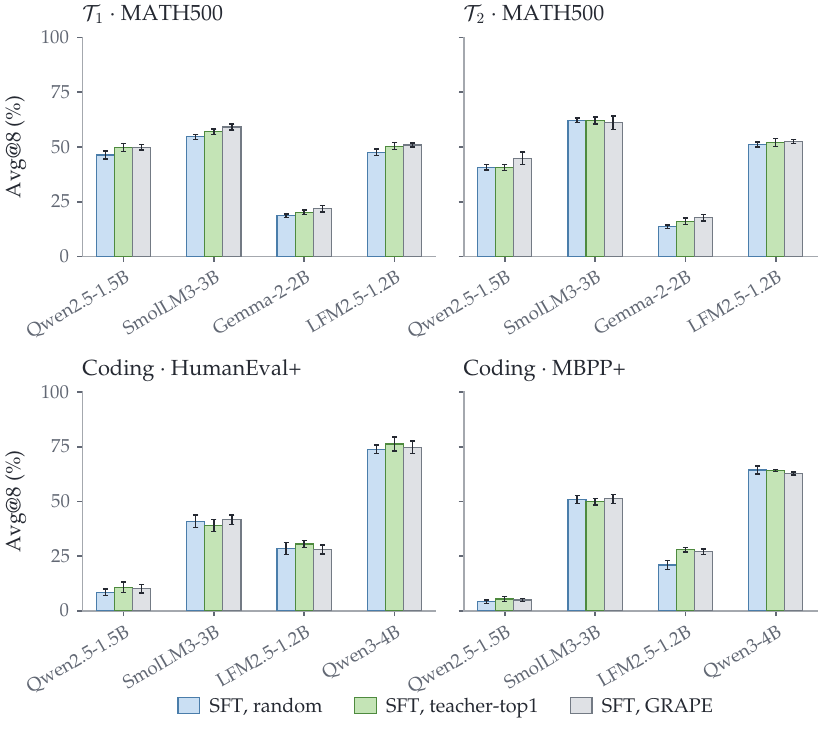}
\caption{\textbf{Random, teacher-top1 and GRAPE selection.} SFT on one correct rollout per solvable prompt, chosen at random, by saved teacher likelihood or by base-student likelihood on MATH500 for $\mathcal{T}_1$ and $\mathcal{T}_2$ (top, the four non-Qwen3 students) and on HumanEval+ and MBPP+ in coding (bottom).}
\label{fig:grape}\label{fig:math-selector-revised}
\end{figure}

\begin{figure}[htbp]
\centering
\includegraphics[width=0.88\linewidth]{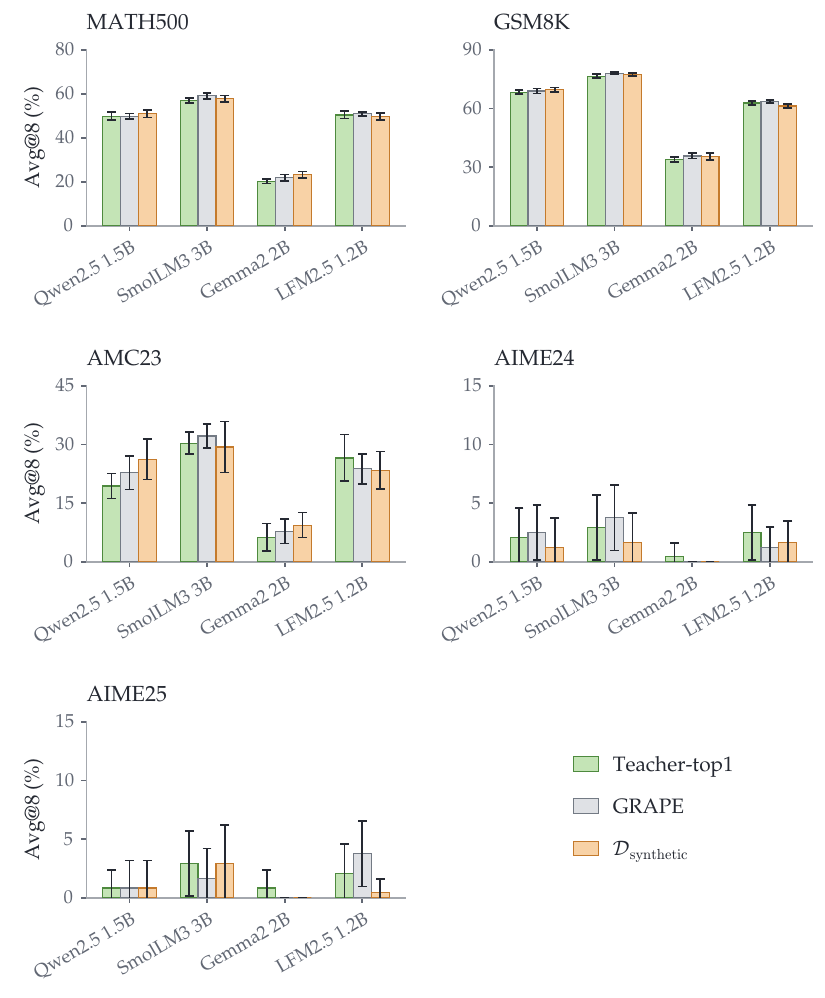}
\caption{\textbf{Teacher-top1 and GRAPE across all maths benchmarks ($\mathcal{T}_1$).} Avg@8 (\%) of SFT on one correct rollout per solvable prompt, chosen by saved teacher likelihood (teacher-top1) or by base-student likelihood (GRAPE), with \Dsynth{} for reference.}
\label{fig:selector-crossbench-t1}
\end{figure}

\begin{figure}[htbp]
\centering
\includegraphics[width=0.88\linewidth]{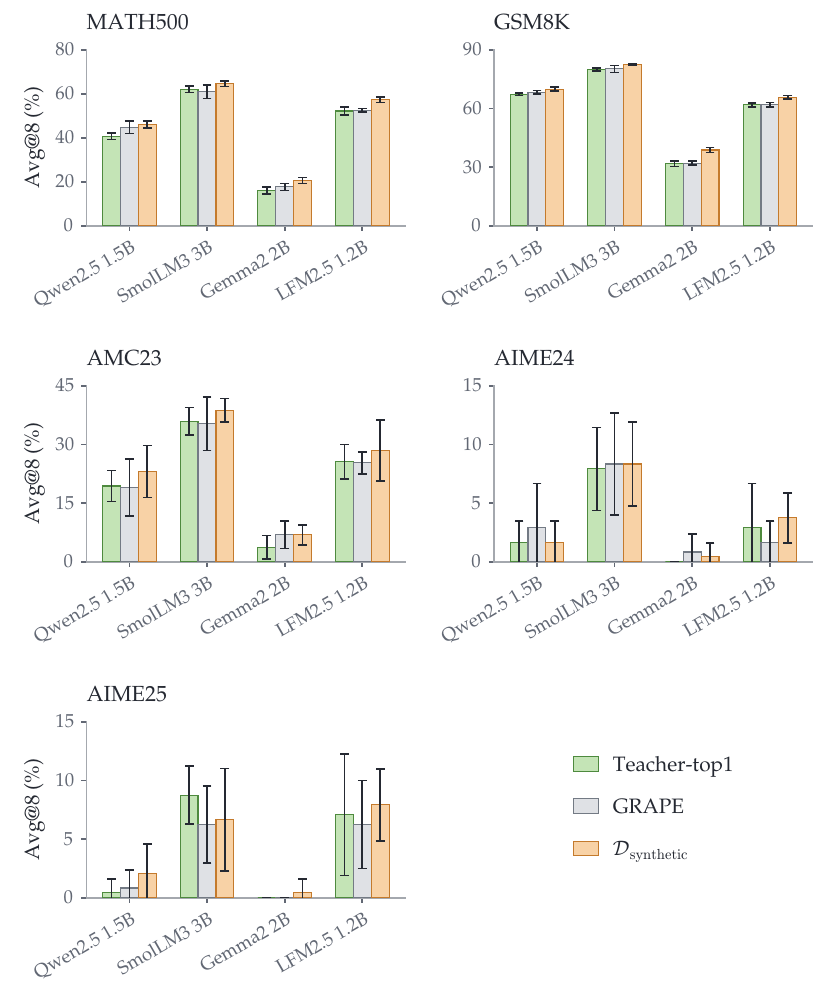}
\caption{\textbf{Teacher-top1 and GRAPE across all maths benchmarks ($\mathcal{T}_2$).} The same comparison for $\mathcal{T}_2$, whose rollouts come from DeepScaler while \Dsynth{} regenerates on Math12K.}
\label{fig:selector-crossbench-t2}
\end{figure}

Figure~\ref{fig:grape} compares single-stage SFT on one correct rollout per solvable prompt selected at random (\DsftRand{}), by cached teacher log-likelihood (teacher-top1), or by base-student log-likelihood (GRAPE~\citep{zhang2026the}) across identical solvable prompts. In maths, GRAPE is $1.05$ and $1.23$ points above teacher-top1 on MATH500 for $\mathcal{T}_1$ and $\mathcal{T}_2$, and $0.61$ and $0.34$ points above it averaged across all five maths benchmarks (Figures~\ref{fig:selector-crossbench-t1} and~\ref{fig:selector-crossbench-t2}). In coding, across the 3,252 solvable tasks, teacher-top1 is $0.43$ points above GRAPE on average (95\% CI $[-0.08,+0.97]$).

\FloatBarrier
\subsection{Which RL Steps the Selected Responses Come From}
\label{app:temporal-provenance}
\begin{figure}[H] %
\centering
\includegraphics[width=0.95\linewidth]{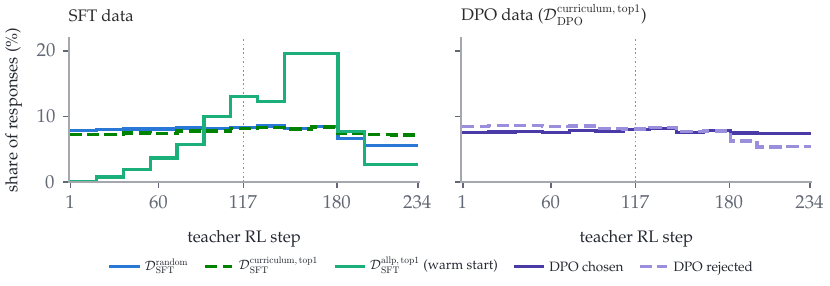}
\caption{\textbf{Teacher RL steps of the selected $\mathcal{T}_1$ responses.} Share of responses per 18-step bin for the SFT datasets (left) and for the chosen and rejected responses of \DdpoCurr{} (right); the dotted line marks the middle of training (step 117). The warm start (\DsftAllp{}) draws mostly from the second half of training, while the DPO pairs span the whole run, with rejected responses slightly earlier on average.}
\label{fig:step-provenance}
\end{figure}

Most $\mathcal{T}_1$ prompts are seen at a single RL step, so the selection rule usually chooses among the 16 rollouts of one step. Even so, our recipe leans towards later teacher checkpoints in two ways (Table~\ref{tab:step-provenance} and Figure~\ref{fig:step-provenance}). First, the warm start (\DsftAllp{}) takes most of its responses from the second half of training, since all-correct prompts ($\mathbb{E}[r](x)=1$) become more common as the teacher improves. Second, for the 2,859 prompts seen at more than one step, teacher-top1 picks a response from the latest step 72\% of the time (compared with 57\% for random selection).

\begin{table}[htbp]
\centering
\small
\begin{tabular}{@{}lrrrr@{}}
\toprule
Responses & Rows & Mean step & Median step & Steps 1--117 \\
\midrule
\DsftRand{} & 10,897 & 112.1 & 112 & 52.8\% \\
SFT, teacher-top1 (\DsftCurr{}) & 10,897 & 118.1 & 119 & 49.2\% \\
\DsftAllp{} & 519 & 140.1 & 147 & 28.1\% \\
\DdpoCurr{}, chosen & 10,378 & 117.0 & 117 & 50.3\% \\
\DdpoCurr{}, rejected & 10,378 & 108.9 & 107 & 55.0\% \\
\bottomrule
\end{tabular}
\caption{\textbf{Source RL steps of the selected $\mathcal{T}_1$ responses.} The teacher is trained for 234 steps; the last column gives the share of responses from the first half of training.}
\label{tab:step-provenance}
\end{table}

\FloatBarrier

\section{Base-Student Likelihood of the DPO Pairs}
\label{app:support-t2}
\begin{figure}[H]
\centering
\includegraphics[width=0.95\linewidth]{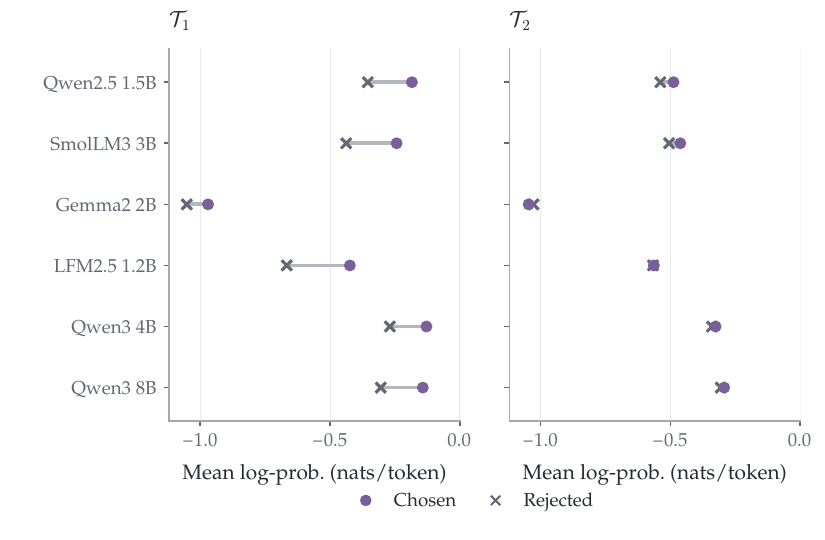}
\caption{\textbf{Base-student log-likelihood of DPO pairs.} Mean base-student log-likelihood (nats/token) of the chosen and rejected DPO responses, normalised by completion length. Chosen responses receive higher likelihood for every student except Gemma-2-2B on $\mathcal{T}_2$; Qwen3-8B-Base, the base model of $\mathcal{T}_2$, is shown for reference. Values use each student's tokenizer.}
\label{fig:chosen-rejected-support}
\end{figure}

Figure~\ref{fig:dpo-vs-sft} in the main text plots the difference between DPO and SFT against the base-student likelihood of the chosen responses. Figure~\ref{fig:chosen-rejected-support} compares the chosen and rejected responses in maths: the chosen responses are more likely than the rejected ones for almost every student, and Gemma-2-2B assigns both the lowest likelihood.

\FloatBarrier
\section{Assets and Licenses}
\label{app:licenses}
We use every asset under its license or terms of use, and the benchmarks for evaluation only.
\begin{itemize}[leftmargin=1.5em,itemsep=1pt,topsep=2pt]
    \item \textbf{Models.} Qwen2.5-1.5B~\citep{qwen2.5}, Qwen3-4B-Base and Qwen3-8B-Base~\citep{yang2025qwen3technicalreport}: Apache-2.0. SmolLM3-3B-Base~\citep{bakouch2025smollm3}: Apache-2.0. Gemma-2-2B~\citep{gemmateam2024gemma2}: Gemma Terms of Use. LFM2.5-1.2B-Base~\citep{liquidai2025lfm2}: LFM Open License v1.0.
    \item \textbf{Training data.} MATH and Math12K~\citep{hendrycks2021measuringmathematicalproblemsolving}: MIT. DeepScaler~\citep{deepscaler2025}: MIT. AceCode-87K~\citep{zeng2025acecoderacingcoderrl}: MIT.
    \item \textbf{Evaluation benchmarks.} MATH500 (PRM800K split)~\citep{lightman2023letsverifystepstep}: MIT. GSM8K~\citep{cobbe2021training}: MIT. HumanEval+ and MBPP+ (EvalPlus)~\citep{liu2023evalplus}: Apache-2.0. AIME 2024, AIME 2025 and AMC 2023: problems published by the Mathematical Association of America~\citep{maa_competitions}.
    \item \textbf{Software.} verl~\citep{sheng2024hybridflow}, TRL~\citep{vonwerra2020trl}, vLLM~\citep{kwon2023efficient}, DeepSpeed, Math-Verify~\citep{kydlicek2024mathverify} and SandboxFusion~\citep{bytedanceseedfoundationcodeteam2025fullstackbenchevaluatingllms}: Apache-2.0.
\end{itemize}

\section{Qualitative Examples from Teacher Training}
\label{app:qualitative}

Saved responses vary both across teacher checkpoints and among attempts at the same question. Below, we compare responses retained by random and teacher-top1 selection, examine how solutions evolve across teacher RL checkpoints, and highlight a failure mode where likelihood selection favours repetitive text.

\subsection{Correct Responses Can Supply Different Training Targets}
\label{app:qualitative-selection}

Correctness filtering alone does not remove all undesirable content from a completion. In coding, the verifier extracts a Python code block for execution, whereas SFT trains the student on the full completion. A correct program followed by unrelated continuation text can therefore pass the verifier while still introducing noisy supervision during SFT.

Figure~\ref{fig:qualitative-archive-code} shows such an example on \texttt{unique\_elements}, a task requiring deduplication while preserving first-occurrence order. At teacher step 10, four of the 16 rollouts pass the verifier. Random selection (\DsftRand{}) picks a 6,945-token rollout that gives a correct implementation and then continues into unrelated Android permission code. From the same group, teacher-top1 selects a clean 114-token solution (which is also used as the chosen response in \DdpoCurr{}). We see similar cases in maths: for a question asking what fraction of a wall Heidi can paint in 9 minutes if the full wall takes 45 minutes, random selection picks a step-15 rollout that reaches $1/5$ but continues into an unrelated spelling discussion, whereas teacher-top1 picks a concise rate calculation from step 225.

\begin{figure}[htbp]
\small
\textbf{Task.} Implement \texttt{unique\_elements} to preserve the order
of first appearances. Both selected functions maintain a set
\texttt{seen}; the excerpts show their final loops and subsequent text.
\par\medskip\hrule\medskip
\begin{minipage}[t]{0.47\linewidth}
\textbf{Random selection, step 10}\par
\textbf{Correct; 6,945 tokens}\par\smallskip
\begin{verbatim}
    for num in arr:
        if num not in seen:
            seen.add(num)
            result.append(num)
    return result
\end{verbatim}
\textit{The Python block ends. Unrelated text follows, including:}\par
\begin{verbatim}
 ActivityCompat.requestPermissions(
 activity,
 arrayOf(permission),
 PERMISSIONS_REQUEST_CODE
 )
\end{verbatim}
\end{minipage}\hfill
\begin{minipage}[t]{0.47\linewidth}
\textbf{Teacher-top1, same step 10}\par
\textbf{Correct; 114 tokens}\par\smallskip
\begin{verbatim}
    for num in arr:
        if num not in seen:
            seen.add(num)
            unique.append(num)
    return unique
\end{verbatim}
\textit{A short usage example follows; no unrelated continuation.}
\end{minipage}
\par\medskip\hrule
\caption{\textbf{Correctness filtering retains an unrelated continuation under random selection.}
The random and teacher-top1 responses are selected from the same step-10 GRPO group. Both pass unit tests, but random selection retains 6,945 tokens with unrelated Android boilerplate outside the Python block, whereas teacher-top1 selects a concise 114-token solution.}
\label{fig:qualitative-archive-code}
\end{figure}

\subsection{Changes in Solutions during Teacher RL}
\label{app:qualitative-evolution}

Figures~\ref{fig:qualitative-math-evolution} and~\ref{fig:qualitative-code-evolution} illustrate how teacher solutions improve over the course of GRPO training, showing why pooling rollouts naively mixes early exploratory errors with cleaner later solutions.

In maths (Figure~\ref{fig:qualitative-math-evolution}), an early $\mathcal{T}_1$ checkpoint (step 55) acknowledges that the boxes are indistinguishable but applies the stars-and-bars formula as though they were labelled ($\binom{6}{1}=6$), whereas the late checkpoint (step 230) correctly counts the three unordered partitions of $5$ into at most $2$ parts. In coding (Figure~\ref{fig:qualitative-code-evolution}, \texttt{Mbpp/167}), the prompt explicitly requires \texttt{next\_power\_of\_2(0) == 1}; the base model nevertheless applies \texttt{math.log2} at zero, whereas the step-120 teacher handles the zero boundary case and uses integer doubling.

\begin{figure}[htbp]
\small
\textbf{Prompt.} How many ways are there to put 5 balls in 2 boxes if the
balls are not distinguishable and neither are the boxes?
\par\medskip\hrule\medskip
\begin{minipage}[t]{0.47\linewidth}
\textbf{Step 55: incorrect}\par\smallskip
\ldots\ This is a classic problem of distributing indistinguishable objects
(balls) into distinguishable bins (boxes).\par\smallskip
\ldots\ In our case, n = 5 (balls) and k = 2 (boxes). So, we have:\par
\(C(5 + 2 - 1, 2 - 1) = C(6, 1)\)\par\smallskip
\ldots\ \(\boxed{6}\)
\end{minipage}\hfill
\begin{minipage}[t]{0.47\linewidth}
\textbf{Step 230: correct}\par\smallskip
\ldots\ The partitions of 5 into up to 2 parts are:
\begin{enumerate}[leftmargin=1.6em,itemsep=1pt,topsep=3pt]
\item 5 (all balls in one box)
\item 4 + 1 (4 balls in one box and 1 ball in the other)
\item 3 + 2 (3 balls in one box and 2 balls in the other)
\end{enumerate}
\ldots\ Since the boxes are indistinguishable, the order of the parts does
not matter\ \ldots\par\smallskip
The final answer is \(\boxed{3}\).
\end{minipage}
\par\medskip\hrule
\caption{\textbf{The later $\mathcal{T}_1$ checkpoint correctly counts indistinguishable boxes.}
The step-55 response counts allocations to labelled boxes; the step-230 response counts unordered partitions.}
\label{fig:qualitative-math-evolution}
\end{figure}

\begin{figure}[htbp]
\small
\textbf{Prompt.} Write a python function to find the smallest power of 2
greater than or equal to n.\par
\texttt{assert next\_power\_of\_2(0) == 1}
\par\medskip\hrule\medskip
\begin{minipage}[t]{0.47\linewidth}
\textbf{Step 0: fails execution tests}\par\smallskip
\begin{verbatim}
import math

def next_power_of_2(n):
    return 2 ** math.ceil(math.log2(n))

assert next_power_of_2(0) == 1
\end{verbatim}
\end{minipage}\hfill
\begin{minipage}[t]{0.47\linewidth}
\textbf{Step 120: passes execution tests}\par\smallskip
\begin{verbatim}
def next_power_of_2(n):
    if n == 0:
        return 1
    power = 1
    while power < n:
        power *= 2
    return power
\end{verbatim}
\end{minipage}
\par\medskip\hrule
\caption{\textbf{Later coding checkpoint handles an explicit zero-input boundary case.}
On \texttt{Mbpp/167}, the step-0 program applies a logarithm at zero; the step-120 program returns one for $n=0$ and uses integer doubling.}
\label{fig:qualitative-code-evolution}
\end{figure}

\subsection{A Limitation of Likelihood-Based Selection}
\label{app:qualitative-selector-limit}

Saved likelihood is not a universal measure of response quality. At coding step 39, the task \texttt{employee\_description} asks for a formatted name and age. Teacher-top1 selects a correct program followed by repetitions of \texttt{assistant} up to the 8,192-token generation cap, whereas the random selection is 37 tokens long. Both pass the verifier, but the repetitive response has a higher saved mean log-likelihood because repeated tokens incur near-zero loss. Across all 3,252 solvable coding groups, teacher-top1 lowers median response length from 100 to 79 tokens relative to random selection, but selects 34 responses that hit the maximum length of 8,192 tokens (versus 14 for random selection). Pairing likelihood ranking with a simple length or repetition filter would be a useful practical safeguard.

\par

\par

\clearpage
\section*{NeurIPS Paper Checklist}

\begin{enumerate}

\item {\bf Claims}
    \item[] Question: Do the main claims made in the abstract and introduction accurately reflect the paper's contributions and scope?
    \item[] Answer: \answerYes{}.
    \item[] Justification: {The abstract and introduction claim that curated reuse of saved teacher RL rollouts achieves comparable average performance to canonical distillation without additional generation, not that it wins in every setting. Averaged over two maths teachers ($\mathcal{T}_1$ in-distribution, $\mathcal{T}_2$ cross-distribution), five students and five benchmarks, the two-stage recipe is 0.37 points below \Dsynth{} (95\% CI $[-0.82,+0.08]$), compared with 1.95 points below for naive rollout SFT; in Python coding, it is 0.89 points above \Dsynth{} over four students on HumanEval+ and MBPP+ (Section~\ref{sec:q1}; Appendix~\ref{app:full-results}). Section~\ref{sec:results} states where the recipe falls short, most clearly for Gemma-2-2B, and the limitations are discussed in Section~\ref{sec:discussion}.}

\item {\bf Limitations}
    \item[] Question: Does the paper discuss the limitations of the work performed by the authors?
    \item[] Answer: \answerYes{}.
    \item[] Justification: {The Limitations paragraph of Section~\ref{sec:discussion} covers the scale we study (4B and 8B teachers, 1.2--4B students), the restriction to tasks with binary verifier rewards, and the use of a single training run per setting; Appendix~\ref{app:qualitative-selector-limit} shows that teacher likelihood can prefer repetitive responses.}

\item {\bf Theory assumptions and proofs}
    \item[] Question: For each theoretical result, does the paper provide the full set of assumptions and a complete (and correct) proof?
    \item[] Answer: \answerNA{}.
    \item[] Justification: {The paper is purely empirical; there are no theorems, lemmas, or formal claims that require proof. The only mathematical objects are dataset definitions and the standard GRPO/DPO/SFT objectives, which are stated as setup rather than as results. We make an intuitive assumption that teacher and student likelihoods have some correlation (Section~\ref{sec:proxy}); this is not a formal claim and is supported by the top-$k$ agreement analysis (Figure~\ref{fig:topk-correct-vs-incorrect}).}

\item {\bf Experimental result reproducibility}
    \item[] Question: Does the paper fully disclose all the information needed to reproduce the main experimental results of the paper to the extent that it affects the main claims and/or conclusions of the paper (regardless of whether the code and data are provided or not)?
    \item[] Answer: \answerYes{}.
    \item[] Justification: {Section~\ref{sec:exp} describes the maths and coding teachers, the students, the baselines and the training settings, and Section~\ref{sec:method} and Table~\ref{tab:variants} give the construction rule for every dataset, with Eq.~\eqref{eq:teacher-mean-logp} defining the mean completion-token log-likelihood used for teacher-top1 selection. Appendix~\ref{app:training} lists the GRPO, SFT, DPO and evaluation settings, including the coding data preparation, execution reward and checkpoint selection, and our code is released.}

\item {\bf Open access to data and code}
    \item[] Question: Does the paper provide open access to the data and code, with sufficient instructions to faithfully reproduce the main experimental results, as described in supplemental material?
    \item[] Answer: \answerYes{}.
    \item[] Justification: {At \url{https://github.com/vashistht/rollout_reuse}, we release (a) the training and evaluation code for the GRPO teachers and for the student SFT, DPO and two-stage runs; (b) the saved rollouts of $\mathcal{T}_1$, $\mathcal{T}_2$ and the first 120 steps of the coding teacher, with each rollout's verifier reward, mean log-likelihood under the generating policy, RL step and per-prompt pass rate; and (c) every derived dataset of Table~\ref{tab:variants}, with instructions to rebuild the datasets and reproduce the tables. Section~\ref{sec:exp} and Appendix~\ref{app:training} give the training and evaluation settings.}

\item {\bf Experimental setting/details}
    \item[] Question: Does the paper specify all the training and test details (e.g., data splits, hyperparameters, how they were chosen, type of optimizer) necessary to understand the results?
    \item[] Answer: \answerYes{}.
    \item[] Justification: {Section~\ref{sec:exp-training} and Appendix~\ref{app:training} specify the optimiser (AdamW; DeepSpeed ZeRO-2 for DPO), the learning rates and how we chose them, batch sizes and epochs, checkpoint selection (MATH500 in maths and the AceCode validation split in coding), and the evaluation protocol (avg@8 at $T{=}0.7$ on MATH500, GSM8K, AIME 2024, AIME 2025 and AMC 2023, and on HumanEval+ and MBPP+). Per-student details, such as the vLLM attention backend for Gemma-2-2B, are also listed.}

\item {\bf Experiment statistical significance}
    \item[] Question: Does the paper report error bars suitably and correctly defined or other appropriate information about the statistical significance of the experiments?
    \item[] Answer: \answerYes{}.
    \item[] Justification: {Figures report avg@8 with one standard deviation over the eight decoding samples, and averages across students come with 95\% confidence intervals from a paired bootstrap over test problems (Section~\ref{sec:exp-training}); the appendix tables report standard deviations over the eight decoding samples of the same runs. As Section~\ref{sec:discussion} states, each setting uses a single training run, so the intervals reflect evaluation noise only.}

\item {\bf Experiments compute resources}
    \item[] Question: For each experiment, does the paper provide sufficient information on the computer resources (type of compute workers, memory, time of execution) needed to reproduce the experiments?
    \item[] Answer: \answerYes{}.
    \item[] Justification:

    {The central compute claim concerns generation. Table~\ref{tab:generation-cost} (Section~\ref{sec:results-headline}) reports the rollouts already produced during teacher RL (239,592 completions for $\mathcal{T}_1$ and 61,440 for the coding archive) and the regeneration pass that reuse avoids (96,000 completions and 58.26M tokens in maths for $\mathcal{T}_1$; 163,840 completions and 17.56M tokens in coding, which took 2.01 generation-stage GB200 GPU-hours). Each maths GRPO teacher trained for about four days on eight L40S GPUs (about 770 GPU-hours per teacher), and student training, evaluation and development used two eight-GPU H100 nodes for about ten days (about 3.8k GPU-hours); the coding teacher and students ran on four-GPU GB200 nodes. Scoring the coding archive for GRAPE took 0.11--0.27 allocated GPU-hours per student.}

\item {\bf Code of ethics}
    \item[] Question: Does the research conducted in the paper conform, in every respect, with the NeurIPS Code of Ethics \url{https://neurips.cc/public/EthicsGuidelines}?
    \item[] Answer: \answerYes{}.
    \item[] Justification: {The work uses publicly released base models (Qwen2.5/Qwen3, SmolLM3, Gemma-2, LFM2.5) and publicly released maths datasets and benchmarks (MATH, GSM8K, AIME 2024/2025, AMC 2023, DeepScaler). No human subjects, no private data, no scraped content beyond what is already redistributed by the cited datasets. The public coding assets include AceCode, HumanEval+, and MBPP+; the coding splits are decontaminated and reference solutions are executed before use.}

\item {\bf Broader impacts}
    \item[] Question: Does the paper discuss both potential positive societal impacts and negative societal impacts of the work performed?
    \item[] Answer: \answerYes{}.
    \item[] Justification: {The work targets the \emph{accessibility} of reasoning distillation: the canonical \Dsynth{} pipeline requires a separate post-RL teacher inference pass at scale, which adds teacher inference on top of the RL run. Reusing rollouts that GRPO has already produced cuts that cost to zero and lowers the barrier to building strong small reasoning students. We therefore also view this work as a call for the open community releasing RL post-training artefacts to ship the rollout pool ($x$, completion, verifier reward, mean completion-token behaviour-policy log-likelihood, originating RL step, pooled per-prompt pass rate) alongside the converged checkpoint, since these by-products are the input to our pipeline and are otherwise discarded. On the negative side, cheaper distillation inherits the standard reasoning-LM misuse caveat -- compressing capabilities (including undesirable ones) into smaller, more redistributable models -- on a narrow verifier-graded maths and Python testbed; we do not introduce new misuse vectors beyond what the broader reasoning-LM distillation literature already exposes.}

\item {\bf Safeguards}
    \item[] Question: Does the paper describe safeguards that have been put in place for responsible release of data or models that have a high risk for misuse (e.g., pre-trained language models, image generators, or scraped datasets)?
    \item[] Answer: \answerNA{}.
    \item[] Justification: {We release code and the saved rollouts of maths and coding teachers trained from public base models on public datasets. These are task-specific responses to maths and Python problems, with no scraped data and no general-purpose model, so they pose no high risk of misuse.}

\item {\bf Licenses for existing assets}
    \item[] Question: Are the creators or original owners of assets (e.g., code, data, models), used in the paper, properly credited and are the license and terms of use explicitly mentioned and properly respected?
    \item[] Answer: \answerYes{}.
    \item[] Justification: {Every model, dataset, benchmark and software asset is credited in Appendix~\ref{app:licenses}, and we use each under its license or terms of use, and the benchmarks for evaluation only. Models: Qwen2.5-1.5B, Qwen3-4B-Base, Qwen3-8B-Base and SmolLM3-3B-Base (Apache-2.0), Gemma-2-2B (Gemma Terms of Use) and LFM2.5-1.2B-Base (LFM Open License v1.0). Data: MATH, Math12K and MATH500 (PRM800K), GSM8K, DeepScaler and AceCode-87K (MIT); HumanEval+ and MBPP+ (EvalPlus, Apache-2.0); and AIME and AMC problems published by the MAA. Software: verl, TRL, vLLM, DeepSpeed, Math-Verify and SandboxFusion (Apache-2.0).}

\item {\bf New assets}
    \item[] Question: Are new assets introduced in the paper well documented and is the documentation provided alongside the assets?
    \item[] Answer: \answerYes{}.
    \item[] Justification: {The new assets are the saved rollouts of our three teachers and the datasets derived from them, released with the code. A data card documents each file's fields (prompt, response, verifier reward, mean log-likelihood, RL step and pass rate), the teacher and RL step that generated each rollout, how the derived datasets are built, and the sources and licenses of the underlying models and datasets.}

\item {\bf Crowdsourcing and research with human subjects}
    \item[] Question: For crowdsourcing experiments and research with human subjects, does the paper include the full text of instructions given to participants and screenshots, if applicable, as well as details about compensation (if any)?
    \item[] Answer: \answerNA{}.
    \item[] Justification: {No crowdsourcing and no human-subject experiments. All grading is automated via Math-Verify on closed-form numeric/symbolic ground truths. Python coding uses deterministic execution tests.}

\item {\bf Institutional review board (IRB) approvals or equivalent for research with human subjects}
    \item[] Question: Does the paper describe potential risks incurred by study participants, whether such risks were disclosed to the subjects, and whether Institutional Review Board (IRB) approvals (or an equivalent approval/review based on the requirements of your country or institution) were obtained?
    \item[] Answer: \answerNA{}.
    \item[] Justification: {No human subjects, so IRB review does not apply.}

\item {\bf Declaration of LLM usage}
    \item[] Question: Does the paper describe the usage of LLMs if it is an important, original, or non-standard component of the core methods in this research? Note that if the LLM is used only for writing, editing, or formatting purposes and does \emph{not} impact the core methodology, scientific rigor, or originality of the research, declaration is not required.
    \item[] Answer: \answerYes{}.
    \item[] Justification: {The teachers and students we study are LLMs; Section~\ref{sec:exp} and Appendix~\ref{app:training} describe them. As declared at submission, we used LLM assistants to edit the text, as coding assistants when implementing and running experiments, and to prepare figures; these uses are not part of the method and do not affect its originality.}

\end{enumerate}

\end{document}